%% file: Paper-src/main.tex
\documentclass{article}

\usepackage[numbers,sort&compress]{natbib}
\IfFileExists{neurips_2024.sty}
  {\usepackage[final]{neurips_2024}}
  {\usepackage[margin=1in]{geometry}}

\usepackage[utf8]{inputenc}
\usepackage[T1]{fontenc}
\usepackage{amsmath}
\usepackage{amssymb}
\usepackage{pifont}
\providecommand{\makecell}[1]{\shortstack{#1}}
\usepackage{booktabs}
\usepackage{multirow}
\usepackage{array}
\usepackage{graphicx}
\usepackage{float}
\usepackage{xcolor}
\usepackage{hyperref}
\usepackage[capitalize]{cleveref}
\usepackage{enumitem}
\usepackage[most]{tcolorbox}
\usepackage{longtable}
\usepackage{array}
\usepackage{booktabs}

\definecolor{outlineblue}{HTML}{3D66A5}
\definecolor{genblue}{HTML}{4285F4}
\definecolor{fixorange}{HTML}{F2994A}
\definecolor{optgreen}{HTML}{34A853}
\definecolor{placeholderred}{HTML}{B3261E}

\hypersetup{
  colorlinks=true,
  linkcolor=outlineblue,
  citecolor=outlineblue,
  urlcolor=purple
}

\newcommand{\GameXpert}{\textsc{GameXpert-Bench}}
\newcommand{\GameGen}{\textsc{GameGen}}
\newcommand{\GameFix}{\textsc{GameFix}}
\newcommand{\GameOpt}{\textsc{GameOpt}}

\newcommand{\yes}{\ding{51}}
\newcommand{\no}{\ding{55}}

\newtcolorbox{outlinebox}[1][Writing outline]{
  breakable,
  title=#1,
  colback=outlineblue!4,
  colframe=outlineblue!70!black,
  fonttitle=\bfseries,
  boxrule=0.6pt,
  arc=1.5mm,
  left=2mm,
  right=2mm,
  top=1mm,
  bottom=1mm
}

\title{GameXpert-Bench: How Far Are Coding Agents from
Expert Game Development?}

\author{
Kun Chen$^{*1,3}$ \quad Haorong Hong$^{*2}$ \quad Peizhong Gao$^{*1,4}$ \quad Jianfeng Lin$^{*2,5}$ \quad Tongxu Luo$^{1,6}$ \quad{Yuxuan Xie$^2$} \quad Chenxu Liu$^1$  \quad Jieling He$^2$\\
\quad Zhongyuan Liu$^{2,\dagger}$ \quad Zeno Zeng$^{1,\dagger}$  \\[0.6em]
$^1$Hunyuan Team, Tencent \quad
$^2$Lightspeed Studios, Tencent \\
$^3$MAIS, Institute of Automation, Chinese Academy of Sciences \\
$^4$Tsinghua University \quad
$^5$The Hong Kong University of Science and Technology \\
$^6$The Chinese University of Hong Kong, Shenzhen\\[0.3em]
{\small $^*$Equal contribution. \quad $\dagger$Corresponding authors.}
}

\begin{document}

\maketitle

\begin{figure}[h]
  \centering
  \includegraphics[width=\linewidth]{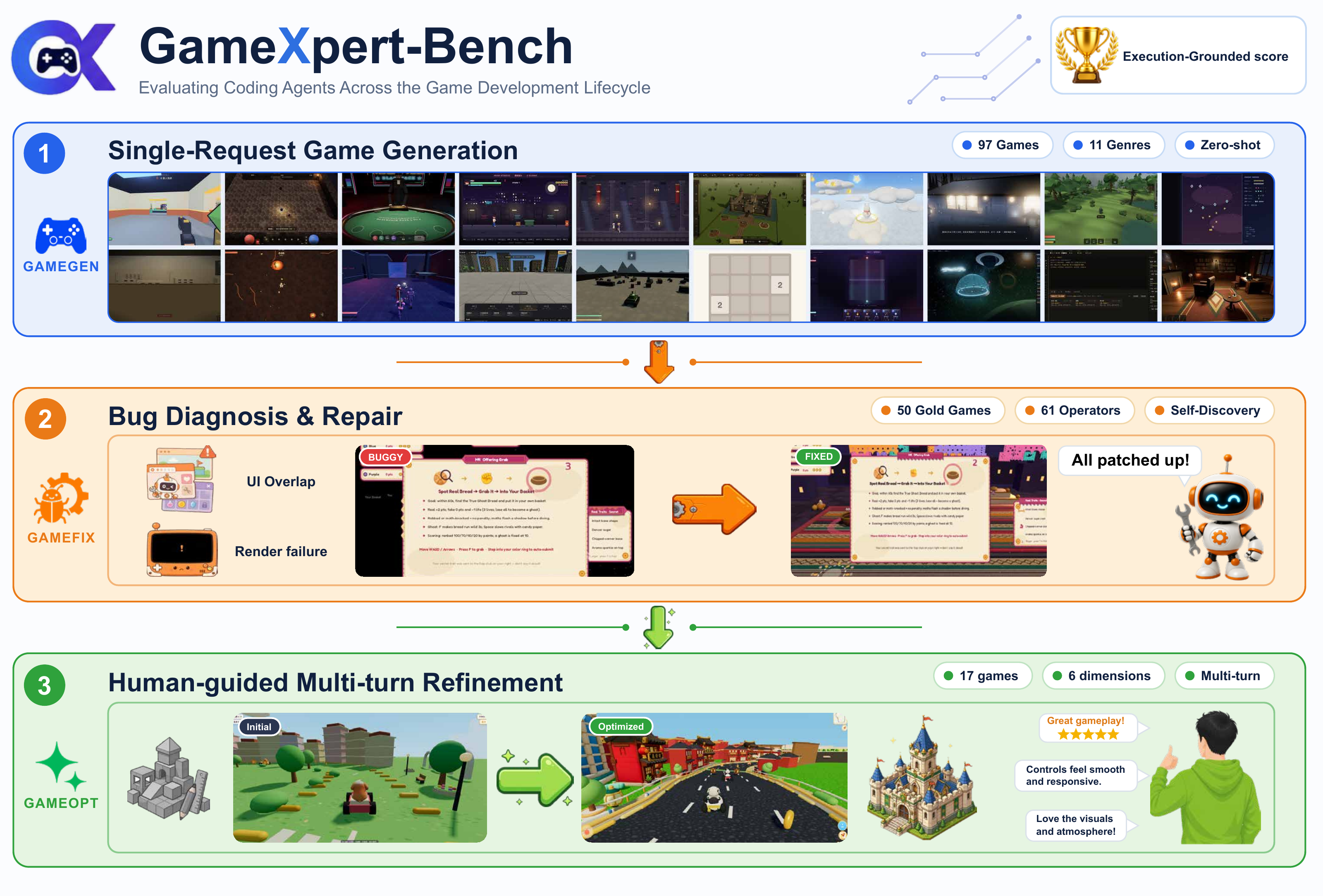}
  \caption{Overview of \GameXpert{} and its three evaluation tracks. GameGen evaluates game creation from natural-language requirements, GameFix evaluates the repair of corrupted reference games, and GameOpt evaluates multi-turn game optimization based on human feedback.}
  \label{fig:banner}
\end{figure}

\begin{abstract}
Recent large language models (LLMs) can operate as coding agents that build complete
games from natural language requests. Game development is especially demanding
because program logic, visual and audio content, interfaces, interaction and
playability must function together in one executable
artifact. Measuring this capability therefore requires evaluation of both game
product and the development process.
Existing benchmarks often assess the game development capabilities of LLMs by evaluating the final artifact or an isolated development stage. Our analysis of complete human--agent development trajectories identifies three stages that together span the lifecycle of game development with a coding agent: initial game generation, bug diagnosis and repair, and optimization over multiple turns.
Therefore, we introduce \GameXpert{}, which operationalizes the three lifecycle stages as three complementary benchmark tracks.  \GameGen{}
evaluates complete game creation from a single request in an empty workspace.
\GameFix{} evaluates diagnosis and repair when defects are reported or left for
the agent to discover. \GameOpt{} evaluates cumulative optimization through
request chains seeded by real development trajectories between users and
agents. We evaluate each track using live game interaction, deterministic
behavioral tests, or final product criteria with regression checks.
The suite contains 97 generation tasks across 11 genres;
100 repair tasks from 50 game levels verified by humans, each with 19--27
injected bugs; and 17 optimization chains with six turns and 102 requests.
Across the three tracks, current agents are more reliable at producing playable foundations and implementing explicit requirements than at discovering defects, verifying runtime behavior, and preserving functionality across changes. This asymmetry shows that initial generation quality alone is insufficient to characterize an agent's game development capability.

\end{abstract}

\input{sections/1_introduction}

\input{sections/2_related_work}
\input{sections/3_overview}
\input{sections/4_gamegen}
\input{sections/5_gamefix}
\input{sections/6_gameopt}
\input{sections/7_conclusion}

\bibliographystyle{unsrt}
\bibliography{ref}

\clearpage
\appendix

\input{appendices/1_gamegen_details}
\input{appendices/2_gamefix_details}
\input{appendices/3_gameopt_details}

\end{document}

%% file: sections/1_introduction.tex
\section{Introduction}
\label{sec:introduction}
The growing ability of large language models (LLMs) to act as coding agents has
made game development an increasingly important testbed for assessing their holistic capability.
Recent systems can implement gameplay features in existing projects and game
engines~\citep{chi2026gamedevbench,la2026gameenginebench} and generate complete
games from natural language specifications
\citep{jiang2026opengame,zhang2026webgamebench,luo2026gamecraft}. Yet successful code generation alone does not yield a playable game. Gameplay logic, rendering, controls, interfaces, audiovisual content, and state transitions must work coherently under player interaction; a failure in any of these elements can degrade the player experience or prevent meaningful play altogether.

This complexity makes evaluation a central challenge. A successful build or a
plausible screenshot does not establish that controls respond correctly,
mechanics remain functional during play, or the completed game satisfies the
request. Recent benchmarks therefore execute generated games and assess their
behavior through browser interaction, gameplay traces, and multimodal judgments
\citep{jiang2026opengame,zhang2026webgamebench,luo2026gamecraft,
jia2026gamegen}. These methods move game evaluation from static code inspection
toward the behavior of the artifact that a player can actually experience. However, evaluating only the final artifact does not reveal how a coding agent's capabilities are exercised throughout the sequence of interactions that produces it. This limitation motivates examining the development process itself, rather than only its endpoint. 
A comprehensive evaluation should therefore consider the complete user-facing game development lifecycle.

We define the user-facing game development lifecycle as the sequence of interactions carried out directly through a coding agent, from the initial request to the final playable artifact. This lifecycle differs in scope from conventional game development frameworks, which include broader organizational phases such as preproduction, production, and postproduction~\citep{aleem2016gamedevlifecycle}.  To derive this lifecycle, we
conduct a qualitative analysis of complete human--agent game development
trajectories. By classifying interactions that change the game according to
their primary intent, we identify three recurring stages: game generation, bug
diagnosis and repair, and optimization over multiple turns. Together, these
stages describe how the executable artifact is created, maintained, and
improved throughout the observed development process.

Despite progress in runtime evaluation, existing benchmarks primarily assess
the final artifact or an isolated stage of development. As summarized in
\Cref{tab:bench-compare}, no existing benchmark suite for games jointly
evaluates all three stages. To address this limitation, we introduce
\GameXpert{}, which connects them within a unified evaluation framework.

\begin{table}[h]\centering\scriptsize
\caption{Coverage of representative benchmarks across the three stages studied
in this paper. Prior work evaluates several constituent capabilities, while
\GameXpert{} places game creation, verified repair, and cumulative optimization
within one benchmark suite. Text labels indicate partial coverage or the source
of feedback.}
\label{tab:bench-compare}
\setlength{\tabcolsep}{4pt}
\renewcommand{\arraystretch}{1.1}
\resizebox{\linewidth}{!}{%
\begin{tabular}{llccc}
\toprule
\textbf{Benchmark} & \textbf{Domain} & \makecell{\textbf{Creation or}\\\textbf{implementation}} & \makecell{\textbf{Diagnosis}\\\textbf{and repair}} & \makecell{\textbf{Iterative}\\\textbf{refinement}} \\
\midrule
GameDevBench~\cite{chi2026gamedevbench}      & Game & \yes & \no & \no \\
GameEngineBench~\cite{la2026gameenginebench} & Game & \yes & \no & \no \\
OpenGame-Bench~\cite{jiang2026opengame}      & Game & \yes & \no & \no \\
WebGameBench~\cite{zhang2026webgamebench}    & Game & \yes & \no & \no \\
GameCraft-Bench~\cite{luo2026gamecraft}      & Game & \yes & \no & \no \\
GBQA~\cite{jiang2026gbqa}                    & Game & \no & Discovery only & \no \\
PlayCoder~\cite{peng2026playcoder}           & GUI apps & Function-level & Explicit issue repair & Agent refiner \\
SWE-Together~\cite{wu2026swe}                & General software & \no & Explicit issue repair & Session simulator \\
\midrule
\multirow{2}{*}{\textbf{GameXpert (ours)}} &
\multirow{2}{*}{Game} &
\multirow{2}{*}{\yes} &
\textbf{Explicit issue +} &
\textbf{Human--agent} \\
& & &
\textbf{self-discovery repair} &
\textbf{co-development} \\
\bottomrule
\end{tabular}%
}
\end{table}

The suite contains one track for each stage. \GameGen{} asks an agent to create
a complete game from a single request
in an empty workspace, without supplied assets or a prescribed engine. It
contains 97 tasks across 11 genres, including 44 3D games. \GameFix{} uses 50
confidential game levels verified by human reviewers and injects 19--27 defects
into each level through reversible mutations. Each level is evaluated with the
defects reported and with selected defects hidden, yielding 100 repair tasks per
run. \GameOpt{} contains 17 optimization chains with six turns each, for a total
of 102 requests. The chains are seeded by development trajectories between
users and agents and completed, where necessary, with requests grounded in the
same game state.

The evaluation protocol for each track reflects its task. \GameGen{} combines
behavioral rubrics, code inspection, live interaction, and human assessment.
\GameFix{} uses deterministic Fail-to-Pass and Pass-to-Pass probes to verify
that a patch repairs the target behavior without introducing regressions.
\GameOpt{} evaluates the final game across gameplay, level design, balance, art,
interface, and audio, while checking whether earlier requested behavior is
preserved. All three protocols follow the same principle: an implementation
receives credit only when the executable game provides evidence of the intended
behavior. Formal task definitions are given in \Cref{sec:overview}.

The three tracks expose related limitations in current coding agents. In
\GameGen{}, agents establish a playable core more reliably than they deliver rich
content, robust interfaces, and fully integrated runtime behavior. In
\GameFix{}, the models separate sharply when defects are hidden, and
near-complete repair remains uncommon when a task contains multiple bugs. On the
evaluated of \GameOpt{}, leading agents often retain requested
functionality across six turns, but preservation of the core game loop and
balanced improvement across product dimensions are not consistent. Across the
three tracks, agents are more reliable at producing playable foundations and
implementing explicit requirements than at discovering defects, verifying
runtime behavior, and preserving functionality across changes. Initial
generation quality alone is therefore insufficient to characterize an agent's
game development capability.

\paragraph{Contributions.}
Our contributions can be summarized as follows:
\begin{itemize}[leftmargin=*,itemsep=2pt,topsep=2pt]
  \item We introduce \GameXpert{}, a benchmark suite that evaluates game
  generation, bug diagnosis and repair, and optimization over multiple turns.
  These stages are derived from complete human--agent development trajectories
  and span the coding-agent game development lifecycle.
  \item We construct 97 diverse generation tasks, 100 controlled repair tasks
  based on confidential Gold Games and reversible mutations, and 17
  optimization chains containing 102 requests. Their evaluation protocols
  connect implementation evidence to executable game behavior.
  \item We evaluate current coding agents across all three tracks and identify
  a common gap between implementing explicit requirements and autonomously
  discovering, verifying, and controlling the effects of changes to a game.
\end{itemize}

%% file: sections/2_related_work.tex
\section{Related Work}\label{sec:related_work}

\paragraph{Game generation and runtime-grounded evaluation.}
Recent benchmarks evaluate coding agents on increasingly realistic
game-development tasks.
GameDevBench~\citep{chi2026gamedevbench} and
GameEngineBench~\citep{la2026gameenginebench} require agents to implement multimodal or runtime behavior within existing game projects.
OpenGame-Bench~\citep{jiang2026opengame} and
WebGameBench~\citep{zhang2026webgamebench} instead evaluate the generation of
complete, browser-native games, whereas
GameCraft-Bench~\citep{luo2026gamecraft} evaluates end-to-end game generation
in the Godot Engine.
Because many game requirements are observable only during execution, these benchmarks complement static inspection with build validation, browser or engine interaction, replayable traces, and multimodal judging.
Most closely related to our evaluation, GameGen-Verifier \citep{jia2026gamegen} decomposes specifications into independently verifiable keypoints and injects runtime states to test them through bounded interactions.
Despite their different environments and protocols, these works primarily
adopt a \emph{single-request, final-artifact} setting.
\GameGen{} retains this setting as the generation track of \GameXpert{}, but
starts from a blank workspace without a provided project.
Its evaluation constructs a shared behavioral rubric through cross-model event analysis, then injects runtime hooks and simulates actions to verify the corresponding game events.

\paragraph{Game bug diagnosis and repair.}
Automated program-repair benchmarks such as
SWE-bench~\citep{jimenez2024swe} evaluate whether agents can resolve reported
issues in existing repositories.
For games, VideoGameQA-Bench~\citep{taesiri2026videogameqa} studies visual
quality assurance from gameplay images and videos, while
GBQA~\citep{jiang2026gbqa} evaluates whether agents can autonomously discover
injected bugs through interactive exploration.
PlayCoder~\citep{peng2026playcoder} additionally combines behavioral GUI
testing with iterative program repair.
These works capture complementary aspects of testing, bug discovery, and repair, but do not jointly provide a verified clean game, a mechanically reversible mutation, its corresponding gold patch, and regression-aware repair tests.
\GameFix{} provides this controlled construction across seven bug dimensions and evaluates agents under both \emph{explicit-issue} and \emph{self-discovery} settings.
Repairs must both resolve the injected failure and preserve previously correct behavior, as measured by Fail-to-Pass and Pass-to-Pass gates.

\paragraph{Human-guided multi-turn refinement.}
Interactive coding benchmarks increasingly expose requirements and corrections
over multiple turns.
SWE-Together~\citep{wu2026swe} reconstructs tasks from real user--agent coding
sessions, but evaluates agents through an anchored, state-conditioned user
simulator.
Within game generation, Play2Code~\citep{huang2026gui} places a coding agent
and a GUI playtester agent in a sustained automated feedback loop.
\GameOpt{} instead constructs multi-turn optimization tasks from real human--agent game-development trajectories.
Experts calibrate each task's starting point, quality constraints, difficulty, and evaluation rubric, after which agents respond to successive optimization requests.
The resulting games are evaluated by an evidence-grounded judge that combines code, image, and rule evidence, targeting improvement beyond repository-level correctness alone.

Taken together, prior work largely studies game generation, bug discovery and repair, and interactive refinement in isolation.
\GameXpert{} formulates them as three controlled, complementary tracks spanning the game-development lifecycle: generation$\rightarrow$repair$\rightarrow$human-guided optimization.

%% file: sections/3_overview.tex
\section{GameXpert-Bench Overview}
\label{sec:overview}

\begin{figure}[t]
  \centering
  \includegraphics[width=\linewidth]{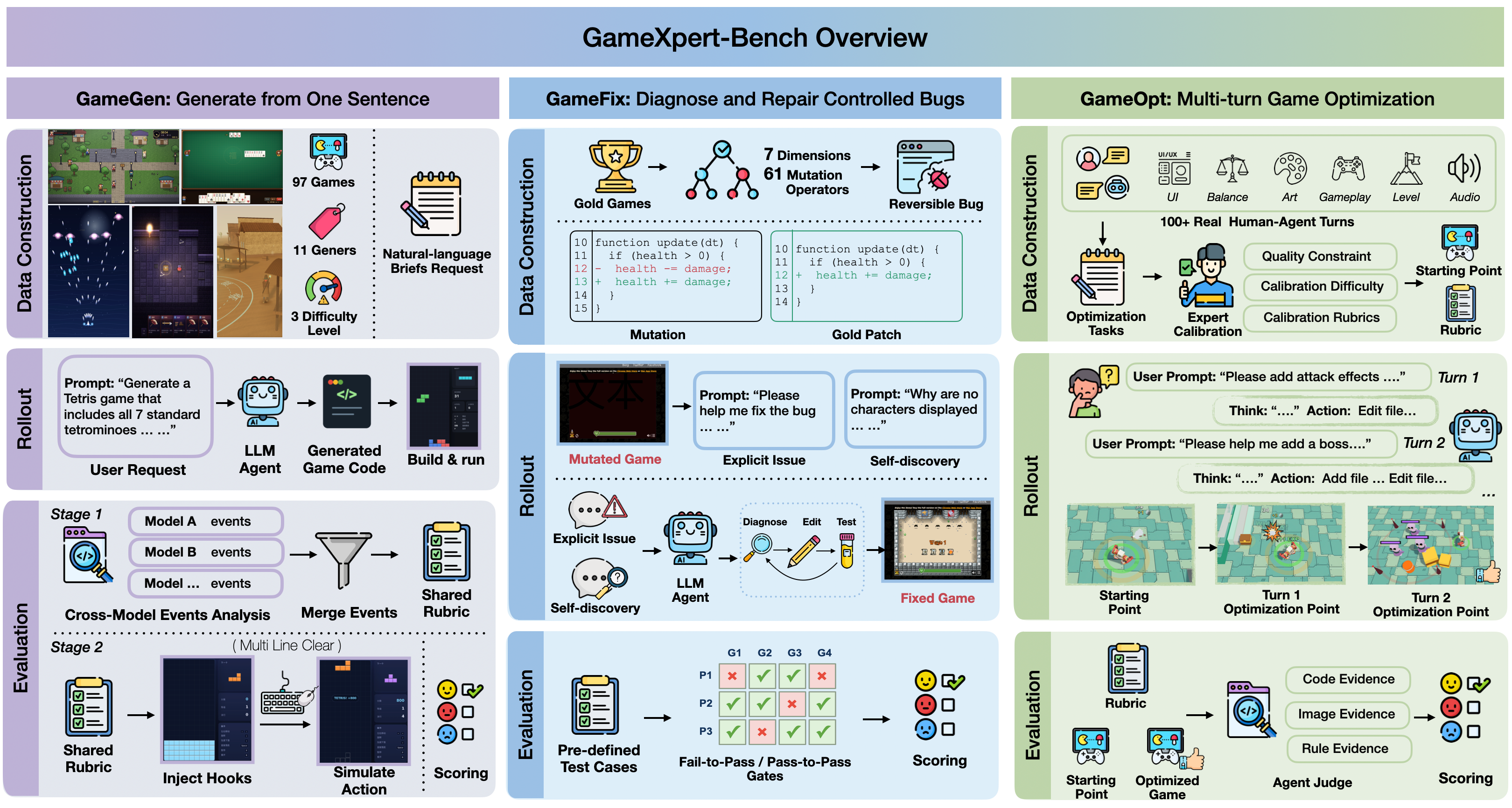}
  \caption{\textbf{The overall structure of \GameXpert{}.} GameGen starts from a natural language request, GameFix starts from a mutated Gold Game, and GameOpt studies multi-turn optimization based on real human interaction.}
  \label{fig:overview}
\end{figure}

\subsection{The Game Development Lifecycle with Coding Agents}

Conventional game development includes broad organizational phases such as
preproduction, production, and postproduction
\citep{aleem2016gamedevlifecycle}. \GameXpert{} focuses on the complete
user-facing artifact lifecycle carried out directly through a coding agent. The
lifecycle begins when a user asks an agent to create a game and ends with the
final playable artifact delivered after requested corrections and improvements.

We derive the lifecycle through a qualitative analysis of complete historical
development trajectories between users and coding agents. We examine each
interaction that creates or modifies the executable game and classify it by its
primary intent through iterative coding. Clarification, environment setup, and
other operational exchanges are associated with the development task they
support but are not treated as separate lifecycle stages. After resolving
ambiguous cases, the analysis identifies three recurring categories that
collectively cover the artifact changes observed across the trajectories.

\begin{outlinebox}[Definition: Complete User-Facing Coding-Agent Game Development Lifecycle]
\[
  \textbf{Generation} \;\rightarrow\; \textbf{Fix} \;\rightarrow\; \textbf{Optimization}.
\]
\textbf{Generation} creates the initial playable game from a user request.
\textbf{Fix} diagnoses and repairs defects revealed in a generated or existing game. \textbf{Optimization} improves a playable game through
successive requests while preserving its established functionality. Together,
the three stages cover the creation, maintenance, and improvement of the
executable artifact.
\end{outlinebox}

\begin{figure}[H]
  \centering
  \includegraphics[width=\linewidth]{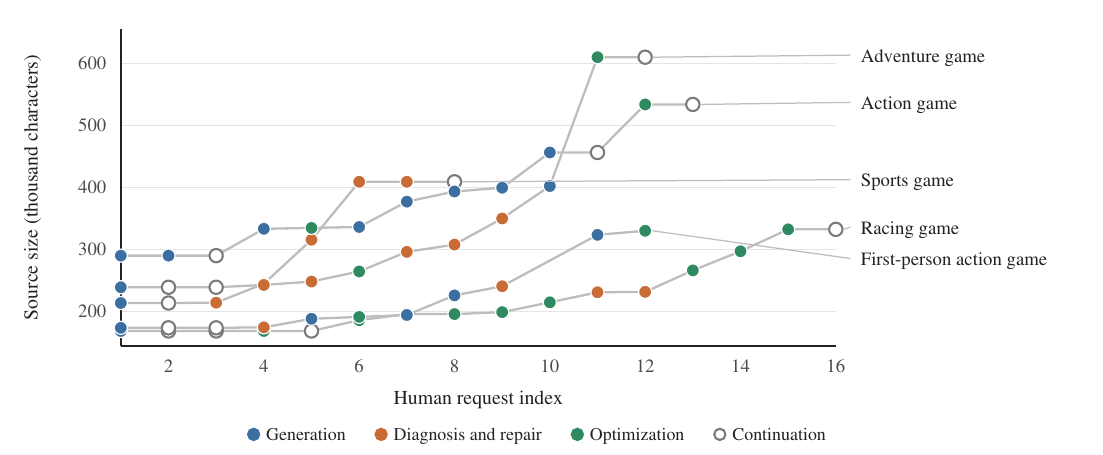}
  \caption{\textbf{Artifact growth across complete coding-agent game
  development trajectories.} Filled markers denote artifact-changing requests
  and colors indicate their lifecycle stage; hollow markers denote continuation
  requests. Each trajectory retains or increases its source size across stages.
  Game names are reported as broad categories.}
  \label{fig:lifecycle-evidence}
\end{figure}

Figure~\ref{fig:lifecycle-evidence} summarizes artifact evolution across the
complete trajectories in our analysis. Source size does not decrease as
generation, diagnosis and repair, and optimization requests operate on
successive versions of each game. This pattern is consistent with later
development work inheriting and extending the results of earlier interactions
rather than beginning from an independent project.

The stages describe recurring modes of development rather than a fixed
pipeline: diagnosis may trigger further repair, and optimization may expose new
defects. For controlled evaluation, \GameXpert{} measures the three stages with
separate, stage-specific corpora:
\GameGen{} evaluates initial creation, \GameFix{} evaluates diagnosis and
repair, and \GameOpt{} evaluates iterative optimization. Together, the tracks
measure whether an agent can create, maintain, and improve an executable game.

\subsection{Three Benchmark Tracks}

\paragraph{\GameGen: Single-request game generation.}
\GameGen{} evaluates whether an agent can turn a natural-language design brief into a complete game. The agent begins in a blank workspace with no provided template, multimedia assets, or prescribed engine, and is therefore responsible for integrating the mechanics, level logic, interactive interface, and audiovisual elements needed for a playable artifact. This track measures the ability to realize a high-level design intent from an initial request.

\paragraph{\GameFix: Game bug diagnosis and repair.}
\GameFix{} evaluates repair in controlled game environments constructed from human-verified Gold Games. Reversible mutation operators inject defects into the source while preserving an exact gold patch and executable behavioral tests. Agents are evaluated both when the reported issues are explicitly identified and when they must discover hidden defects from limited symptom descriptions. A repair succeeds only when it restores the affected behavior and preserves behavior that was correct before the repair.

\paragraph{\GameOpt: Human-guided game optimization.}
\GameOpt{} evaluates multi-turn refinement from a playable game snapshot. Each task replays a sequence of requests derived from human--agent game development trajectories, and the agent continues from the state it produced in the preceding turn. The requests span complementary aspects of game quality, including gameplay, level design, balance, art, interface, and audio. This setting evaluates whether an agent can incorporate successive product-level requests while maintaining the playability and compatibility of the evolving game.

\subsection{Benchmark Organization}

All three tracks are evaluated through the game artifact rather than source code alone. \GameGen{} combines a shared behavioral rubric with runtime verification and human assessment of player-facing qualities. \GameFix{} uses deterministic gameplay probes to verify both repaired and preserved behaviors.
\GameOpt{} uses evidence-grounded criteria over the final game, drawing on code and rendered output as appropriate and accounting for regressions. 
These protocols connect implementation-level changes to the game that players can actually run and experience.

The remainder of the paper presents the three tracks in lifecycle order.
\Cref{sec:gamegen} describes the game-generation corpus and evaluation,
\Cref{sec:gamefix} details the controlled repair setting, and
\Cref{sec:gameopt} introduces the trajectory-based optimization benchmark and its scoring protocol.

%% file: sections/4_gamegen.tex
\section{\GameGen: Single-Request Game Generation}
\label{sec:gamegen}

Within the AI-based game-development lifecycle introduced in
\Cref{sec:overview}, \textbf{GameGen} focuses on its initial generation stage:
translating a high-level design vision into the first playable game artifact.
In this single-turn setting, the coding agent acts as the developer. Given only
one natural-language instruction as the design brief, it must generate a fully
functional, browser-native game.

Importantly, this generation takes place in a blank workspace without provided
game templates or multimedia assets, and the instruction does not prescribe a
particular game engine or development toolchain. The agent is therefore free to
decide whether to use an engine and how to construct the game. This setting
captures a broadly accessible form of game creation through general-purpose
coding agents, while avoiding dependence on any particular engine ecosystem.
It also places greater emphasis on the agent's native coding and game-
engineering capabilities rather than its familiarity with engine-provided
scaffolds. The agent must build and integrate gameplay mechanics, level logic,
interactive UI, and audiovisual content into a cohesive artifact. This setup
makes \textbf{GameGen} a rigorous test of end-to-end game generation. Instead
of merely evaluating the functional correctness of isolated code snippets, our
fine-grained evaluation criteria go beyond whether the game compiles and runs
to assess how faithfully it realizes the intended design.

\subsection{Task Formulation and Game Corpus}
\label{subsec:task-corpus}

\textbf{GameGen} comprises 97 distinct games spanning 11 genres, including 44
tasks that require 3D rendering. Each game is described by a natural-language
brief, and the games are deliberately spread across three difficulty levels,
from those built around a single core mechanic to those that require
coordinating several interacting mechanics. This graded design lets \textbf{GameGen}
measure agents across a wide range of complexity: the easier games
test whether an agent can deliver a coherent and playable build at all, while
the harder ones expose failures that surface only as games grow more complex,
and thus better separate stronger agents from weaker ones.

\paragraph{Formalization.} Let $p_i$ denote the natural-language design brief
for the $i$-th game and $A$ a coding agent. The agent produces a browser-native
artifact $a_i = A(p_i)$ from the brief alone; importantly, $p_i$ does not
prescribe a game engine or implementation stack. The benchmark corpus is the
collection
\begin{equation}
  \mathcal{D} = \{\,p_i\,\}_{i=1}^{N}
\end{equation}
over all $N$ games.

\subsection{Generation Protocol}
\label{subsec:generation-protocol}

Building on the task defined in \Cref{subsec:task-corpus}, \textbf{GameGen} adopts a
strictly from-scratch generation protocol. Each agent is provided solely with
the natural-language brief and is required to synthesize a complete,
browser-native game without being provided with a game template, a prescribed
engine, or pre-existing assets. The agent is free to choose its own technical
stack, including whether to use an engine or general-purpose browser libraries,
and must produce every source file required by the final artifact. Generation
is carried out by a general-purpose coding agent, \textbf{Claude Code}, invoked
directly under its default configuration; we introduce no task-specific
scaffolding, ensuring that the resulting artifact is attributable to the
agent's intrinsic capability rather than to auxiliary engineering. Since the
brief enumerates the game's required features explicitly, the protocol further
probes the agent's \textbf{instruction-following} ability: the agent is expected
to faithfully realize the specified requirements while retaining the latitude
to introduce additional mechanics and refinements. Each task is completed
within a single generation session and yields a self-contained game that
executes directly in the browser.

\subsection{Evaluation Framework}
\label{subsec:eval-framework}

Our evaluation begins only after all evaluated models have completed generation.
We first construct a single \textbf{Shared Rubric} for each game through
\textbf{Cross-Model Events Analysis}. An event-analysis agent examines the
games produced by every model, extracts the gameplay events realized in each
artifact, and preliminarily categorizes them as either core or bonus events.
For each game, we pool the events observed across models, merge semantically
equivalent descriptions, and remove duplicates. Human annotators then review
the resulting candidate pool and curate a unified checklist, using the observed
cross-model coverage of each event as an important reference. The resulting
Shared Rubric is denoted as
\begin{equation}
  \mathcal{C}_i = \mathcal{R}_i \,\uplus\, \mathcal{B}_i,
\end{equation}
where core events $\mathcal{R}_i$ capture the relatively stable behaviors
necessary to realize game $i$, while bonus events $\mathcal{B}_i$ capture
additional mechanics and content that vary across models. Aggregating bonus
events across models allows the rubric to distinguish models with different
levels of capability without tailoring the evaluation to any single model.
Every artifact generated for the same game is subsequently assessed against
the same $\mathcal{C}_i$, ensuring a consistent comparison.

Our evaluation dimensions follow the classical decomposition of a game into
its mechanical systems, the resulting gameplay, and the player's
experience~\citep{zubek2020elements}. We assess each generated artifact across
four complementary dimensions.

\paragraph{Completeness (Automated).} Completeness measures the fulfillment of
the core events in $\mathcal{R}_i$, including the fundamental mechanics,
interactive UI behaviors, and level logic required by the game. It therefore
captures whether the generated artifact faithfully implements the essential
content of the design brief rather than merely compiling into an executable
page.

\paragraph{Richness (Automated).} Richness measures the realization of bonus
events in $\mathcal{B}_i$. It rewards supplementary mechanics, interactive
content, level variety, and other functional extensions beyond the core game,
thereby distinguishing structurally elaborate games from minimally viable
ones.

\paragraph{Visual Quality (Human).} Visual quality targets aesthetic and
spatial properties that require human judgment. In addition to overall visual
presentation, annotators assess whether on-screen and in-world elements are
positioned coherently. This includes checking for UI overlaps and occlusions,
content overflowing its designated boundaries, and, in 3D environments,
geometric interpenetration between objects. Such defects directly degrade the
player-facing quality of an otherwise functional artifact.

\paragraph{Player Experience (Human).} Player experience evaluates the
holistic playability of the generated game, including the responsiveness of
interaction flows and the moment-to-moment feel of engaging with the game. It
complements event-level functional evaluation by capturing qualities that
emerge only through actual play.

\paragraph{Hybrid Automated Scoring.} To compute completeness and richness, the
evaluation agent combines static source-code analysis with dynamic runtime
validation. It first inspects the generated codebase to locate the programmatic
logic corresponding to each item in the Shared Rubric. It then interacts with
the live game to reach the relevant states, trigger the targeted events, and
verify that the observed runtime behavior matches the criterion. A checklist
item is credited only when its intended effect is validated during execution;
this constraint prevents syntactically plausible yet functionally inert code
from yielding false-positive scores. The criterion-level outcomes are then
aggregated into completeness and richness scores and combined with the
human-assessed dimensions to provide a holistic evaluation of the generated
artifact. All four dimension scores are normalized to $[0,100]$ and weighted
equally. Let $S_d(A)$ denote the benchmark-level score of model $A$ on
dimension $d$, averaged over all $N$ games. The overall score is computed as
the unweighted mean
\begin{equation}
  \begin{split}
    S_{\mathrm{overall}}(A) = \frac{1}{4}\bigl(&
      S_{\mathrm{comp}}(A) + S_{\mathrm{rich}}(A) \\
      &+ S_{\mathrm{exp}}(A) + S_{\mathrm{vis}}(A)\bigr).
  \end{split}
  \label{eq:gamegen-overall-score}
\end{equation}

\subsection{Experimental Results and Analysis}
\label{subsec:gamegen-results}

\paragraph{Experimental Setup.}
We evaluate 15 representative model variants: Claude-Opus-5~\cite{claude_models}, Claude-Fable-5~\cite{claude_models}, Claude-Opus-4.8~\cite{claude_models}, Claude-Opus-4.7~\cite{claude_models}, Kimi-K3~\cite{kimi_k3}, GPT5.6-sol~\cite{gpt_models}, GPT5.5~\cite{gpt_models}, GLM5.2~\cite{glm_models}, GLM5.1~\cite{glm_models}, DeepSeek-V4-Flash~\cite{deepseek_v4_flash}, Hy3~\cite{hunyuan_3}, Gemini-3.5-flash~\cite{gemini_3_5_flash}, MiniMax-M3~\cite{minimax_m3}, Qwen3.7-Max~\cite{qwen_3_7_max}, and Seed-2.1-pro~\cite{seed_2_1_pro}.
Every model is evaluated on all 97 game briefs under the same
generation protocol described in \Cref{subsec:generation-protocol}: the model receives
the same blank workspace and natural-language brief, with no provided template,
assets, or prescribed game engine. We report scores aggregated over the full
benchmark. The report-level diagnostic analyses below cover all 15 models and
97 games, totaling $15\times97=1{,}455$ model--game runs and 43,081 event-level
outcomes.

\begin{table*}[t]
\centering
\caption{Main results on \GameGen{}. All scores are reported out of 100, with the Overall score computed as the average of Completeness, Richness, Player Experience, and Visual Quality. Values are rounded to one decimal place. The best and second-best results in each column are shown in bold and underlined, respectively.}
\label{tab:gamegen-main-results}
\small
\setlength{\tabcolsep}{5.5pt}
\begin{tabular}{clccccc}
\toprule
\# & \textbf{Model} & \textbf{Overall} & \textbf{Completeness} & \textbf{Richness} & \textbf{Experience} & \textbf{Visual} \\
\midrule
1  & Claude-Opus-5     & \textbf{79.7} & \textbf{94.4} & \textbf{72.0} & \underline{72.4} & \textbf{80.4} \\
2  & Claude-Fable-5    & \underline{75.8} & \underline{89.2} & \underline{62.8} & \textbf{74.4} & \underline{76.8} \\
3  & Kimi-K3           & 71.3 & 85.2 & 56.0 & 70.4 & 73.6 \\
4  & Claude-Opus-4.8   & 69.9 & 84.4 & 49.6 & 72.0 & 74.0 \\
5  & GPT5.6-sol        & 63.4 & 79.6 & 42.0 & 63.2 & 68.8 \\
6  & GLM5.2            & 63.4 & 76.8 & 44.8 & 62.4 & 69.6 \\
7  & DeepSeek-V4-Flash & 62.9 & 81.6 & 48.4 & 57.6 & 64.0 \\
8  & Hy3               & 62.1 & 77.2 & 41.6 & 62.0 & 68.0 \\
9  & Claude-Opus-4.7   & 62.0 & 76.0 & 42.8 & 61.6 & 67.6 \\
10 & MiniMax-M3        & 59.6 & 73.2 & 40.8 & 56.4 & 68.0 \\
11 & Gemini-3.5-flash  & 59.1 & 73.2 & 46.0 & 53.2 & 64.0 \\
12 & GPT5.5            & 59.0 & 74.0 & 36.8 & 60.4 & 65.2 \\
13 & GLM5.1            & 58.0 & 72.8 & 43.2 & 56.4 & 60.0 \\
14 & Qwen3.7-Max       & 54.4 & 66.4 & 34.0 & 55.2 & 61.6 \\
15 & Seed-2.1-pro      & 48.7 & 58.4 & 30.8 & 44.8 & 60.8 \\
\bottomrule
\end{tabular}
\end{table*}

\paragraph{Overall Performance.}
\Cref{tab:gamegen-main-results} presents the main \GameGen{} leaderboard.
Claude-Opus-5 ranks first with an overall score of 79.7, exceeding
Claude-Fable-5 by 3.9 points and Kimi-K3 by 8.4 points. It leads in
completeness (94.4), richness (72.0), and visual quality (80.4), while
Claude-Fable-5 achieves the best player experience (74.4). Thus,
Claude-Opus-5 combines the broadest functional coverage with the strongest
visual score, whereas Claude-Fable-5 retains an advantage in moment-to-moment
player experience. Kimi-K3 and Claude-Opus-4.8 follow with 71.3 and 69.9
overall, respectively. A further 6.5-point drop separates Claude-Opus-4.8 from
GPT5.6-sol, after which GPT5.6-sol, GLM5.2, DeepSeek-V4-Flash, Hy3, and
Claude-Opus-4.7 form a dense middle group between 62.0 and 63.4.

\paragraph{Dimension-wise Performance.}
Across the 15 evaluated models, completeness averages 77.5, substantially
higher than richness at 46.1; player experience and visual quality average
61.5 and 68.2, respectively. Richness is lower than completeness for every
model, revealing a persistent gap between constructing the essential playable
core and extending it with diverse bonus mechanics and content. The leading
models also exhibit complementary profiles. Claude-Opus-5 leads three of the
four dimensions, including visual quality, whereas Claude-Fable-5 remains
strongest on player experience (74.4) and ranks second on visual quality
(76.8). These differences show that a single
notion of ``playability'' cannot capture game-generation quality: functional
breadth and player-facing polish remain related but distinct capabilities.

\paragraph{Progress across Model Versions.}
Within model families represented by multiple generations, Claude-Opus-4.8
improves over Claude-Opus-4.7 by 7.9 overall points, and Claude-Opus-5 adds a
further 9.8 points. The latter gain is largest in richness (+22.4), followed by
completeness (+10.0) and visual quality (+6.4), while player experience changes
only slightly (+0.4). This pattern suggests that the principal improvement
from Claude-Opus-4.8 to Claude-Opus-5 lies in broader functional scope,
accompanied by a substantial gain in presentation quality. GLM5.2 similarly
improves over GLM5.1 by 5.4 overall points, with gains across all four
dimensions.

\begin{figure*}[t]
  \centering
  \includegraphics[width=\textwidth]{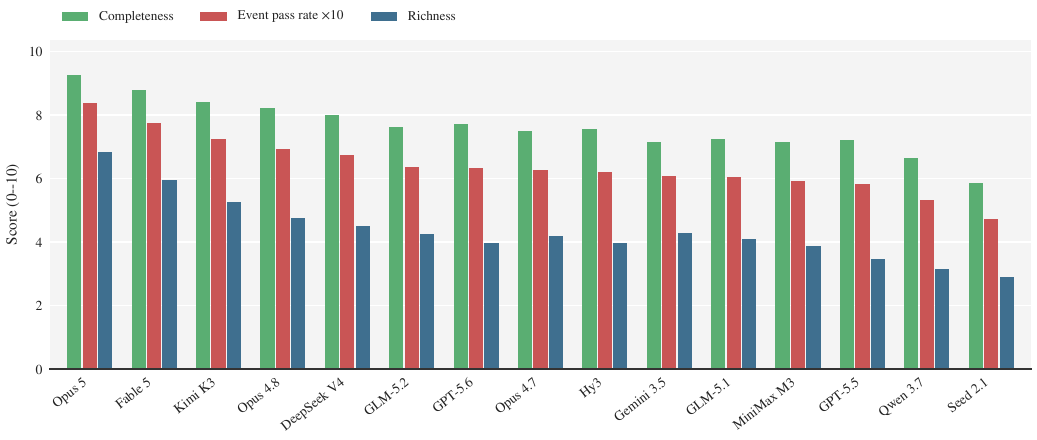}
  \caption{Functional coverage across all 15 models. Average completeness and
  richness are reported on $[0,10]$, and event pass rate is multiplied by 10
  for comparison. Models follow the main-leaderboard order.}
  \label{fig:gamegen-functional-coverage}
\end{figure*}

\begin{figure*}[t]
  \centering
  \includegraphics[width=\textwidth]{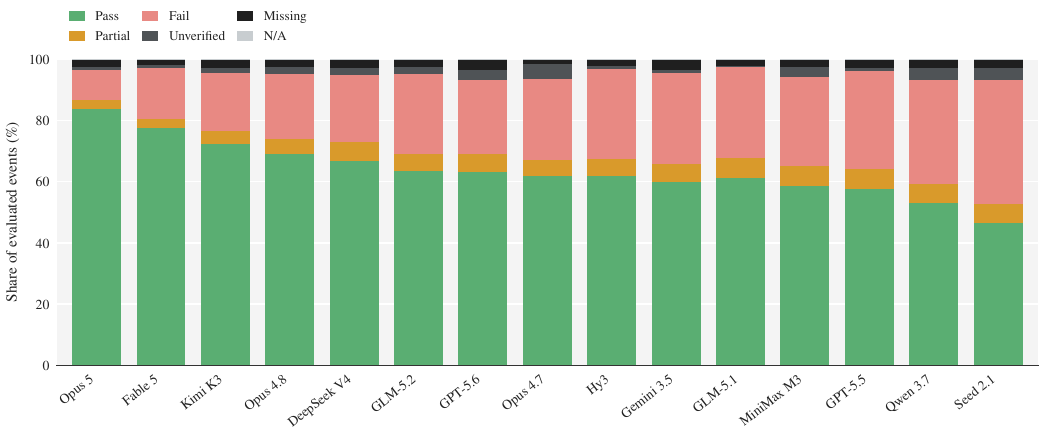}
  \caption{Composition of event-level outcomes across all 15 models, including
  checklist events that are missing from the generated report. Each stacked
  bar sums to 100\%, and models follow the main-leaderboard order.}
  \label{fig:gamegen-event-status}
\end{figure*}

\paragraph{Functional Coverage and Event Outcomes.}
\Cref{fig:gamegen-functional-coverage,fig:gamegen-event-status} provide an
event-level view of the functional scores. Completeness and event pass rate
move together: Claude-Opus-5 passes more than 80\% of tested events, whereas
the lowest-ranked model passes fewer than half. More importantly, richness
remains below completeness for every model, including the leaders. Current
agents therefore tend to prioritize the required gameplay skeleton before
adding optional mechanics, levels, and interactions. The stacked outcome
distribution further shows that the separation between models is driven mainly
by how much probability mass is converted from failed or unresolved events
into fully passed events, rather than by a uniform change in every outcome
category. This result complements the aggregate leaderboard by showing that
stronger models not only attempt broader functionality but also execute a
larger fraction of tested events successfully.

\begin{figure*}[t]
  \centering
  \includegraphics[width=\textwidth]{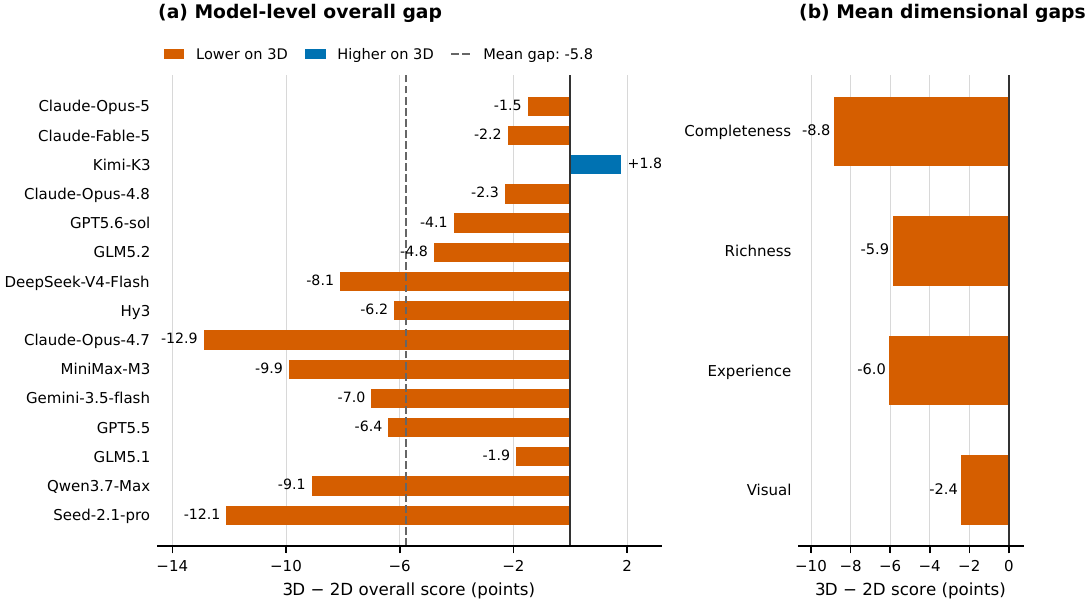}
  \caption{Performance gaps between the 2D and 3D subsets of \GameGen{}.
  \textbf{(a)} Difference in overall score for each model, computed as the 3D
  score minus the 2D score. The dashed line marks the cross-model mean gap.
  \textbf{(b)} Cross-model mean gaps across the four evaluation dimensions.
  All scores are reported on a $[0,100]$ scale, and negative values indicate
  lower performance on the 3D subset. Models follow the main-leaderboard
  order.}
  \label{fig:gamegen-2d3d-gap}
\end{figure*}

\paragraph{2D versus 3D Game Generation.}
We further disaggregate the results by game dimensionality. Averaged across
the 15 evaluated models, the overall score decreases from 65.9 on the 53 2D
games to 60.1 on the 44 3D games, corresponding to an average drop of 5.8
points. As shown in \Cref{fig:gamegen-2d3d-gap}, 14 of the 15 models perform
worse on the 3D subset, with Kimi-K3 being the only exception (+1.8). The
largest mean degradation occurs in completeness ($-8.8$ points), followed by
player experience ($-6.0$), richness ($-5.9$), and visual quality ($-2.4$).
These results indicate that 3D game generation poses a greater challenge to
current agents under our benchmark. Notably, the performance gap is driven
primarily by reduced functional completeness rather than visual quality alone.
Full subset-level leaderboards are provided in the appendix.

\begin{figure*}[t]
  \centering
  \includegraphics[width=\textwidth]{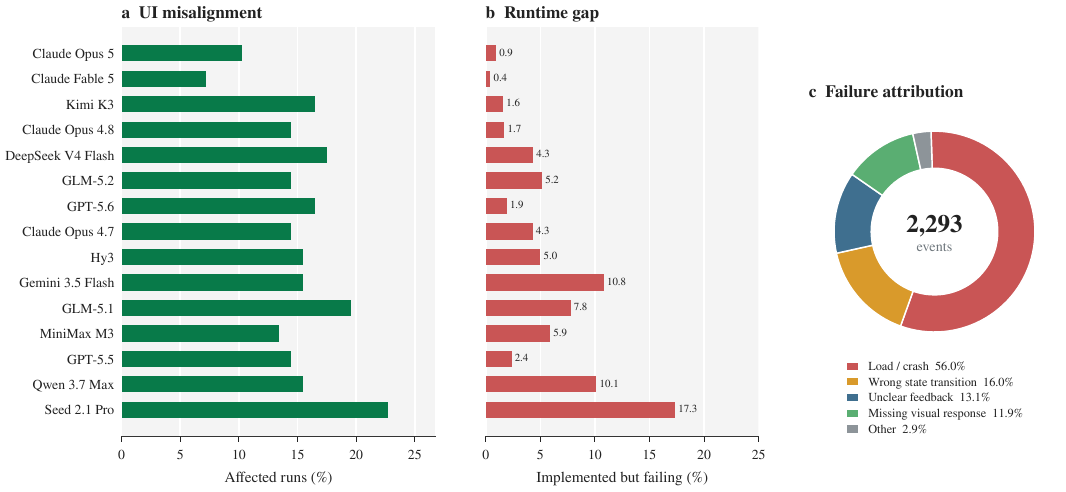}
  \caption{Player-facing and runtime failure diagnostics. \textbf{(a)} Share
  of model--game runs whose evaluation reports contain evidence of UI overlap
  or misalignment. \textbf{(b)} Share of events recognized as implemented in
  code but assigned a runtime outcome of \textsc{Fail} or \textsc{Partial}.
  \textbf{(c)} Attribution of the 2,293 implementation-gap events.}
  \label{fig:gamegen-failure-diagnostics}
\end{figure*}

\paragraph{UI Misalignment Is a Universal Failure Mode.}
We retrieve the \texttt{reasoning} fields of UI-related events from each
evaluation report and search for a fixed multilingual keyword set covering
misalignment, overlap, occlusion, covering, overflow, and offset, together with
the English terms \texttt{overlap}, \texttt{clipped}, \texttt{off-screen}, and
\texttt{overflow}. A model--game run is marked as affected if any of these
terms is matched. As shown in \Cref{fig:gamegen-failure-diagnostics}, 221 of the
1,455 runs (15.2\%) are flagged. Every evaluated model exhibits this problem:
the rate ranges from 7.2\% for Claude-Fable-5 and 10.3\% for Claude-Opus-5 to
19.6\% for GLM5.1 and 22.7\% for Seed-2.1-pro. Although the ordering is not
strictly monotonic, UI layout defects are generally more prevalent among
lower-performing models. These results reveal a widespread gap between
generating individually plausible interface components and composing them into
a stable layout across runtime states. We provide a per-game breakdown in the
appendix.

\paragraph{Implemented Does Not Mean Functional.}
Finally, we examine events for which static inspection reports
\texttt{implementation\_status == implemented}, but dynamic evaluation assigns
\textsc{Fail} or \textsc{Partial}. Such cases correspond to functionality that
appears in the code but cannot be triggered reliably, produces insufficient
feedback, or transitions to an incorrect state at runtime. Across 43,081
assessed event instances, 2,293 (5.32\%) fall into this category. As shown in
\Cref{fig:gamegen-failure-diagnostics}, the rate varies sharply across models,
from 0.4\% for Claude-Fable-5 and 0.9\% for Claude-Opus-5 to 10.1\% for
Qwen3.7-Max, 10.8\% for Gemini-3.5-flash, and 17.3\% for Seed-2.1-pro.

The attribution panel further shows that load or crash failures account for
the majority of these cases (56.0\%). Incorrect state transitions contribute
16.0\%, unclear or missing feedback 13.1\%, and missing visual responses
11.9\%, with the remaining 2.9\% assigned to other causes.
Thus, most implementation gaps arise not because the relevant code is entirely
absent, but because it is not integrated into a robust end-to-end execution
path. This finding directly motivates our hybrid evaluator: static code
inspection is useful for locating candidate implementations, but runtime
interaction is necessary to prevent inert or incorrectly wired functionality
from receiving credit.

\begin{figure*}[t]
  \centering
  \includegraphics[width=0.6\textwidth]{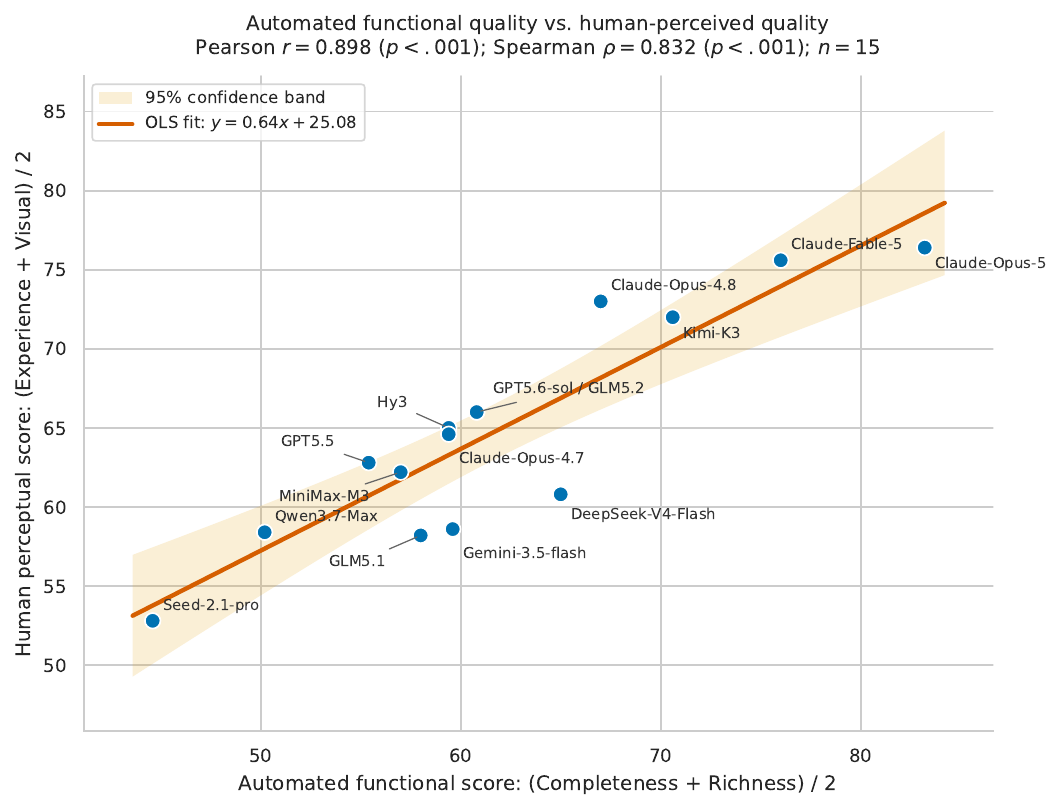}
  \caption{Model-level association between automated functional evaluation and
  human-perceived quality. Each point represents one of the 15 evaluated
  models. The automated functional score is the mean of Completeness and
  Richness, while the human perceptual score is the mean of Player Experience
  and Visual Quality; all component scores are on a $[0,100]$ scale. The solid
  line denotes the ordinary least-squares fit, and the shaded region denotes
  its 95\% confidence band for the mean response. GPT5.6-sol and GLM5.2 have
  identical aggregate coordinates and therefore overlap.}
  \label{fig:gamegen-human-machine-correlation}
\end{figure*}

\paragraph{Automated Functional Scores Are Strongly Associated with
Human-Perceived Quality.}
We examine whether models rated as functionally stronger by the automated
evaluator also tend to produce games that receive higher human judgments. For
each model $A$, we define an automated functional score and a human perceptual
score as
\begin{equation}
  \begin{aligned}
    S_{\mathrm{auto}}(A)
      &=\frac{S_{\mathrm{comp}}(A)+S_{\mathrm{rich}}(A)}{2}, \\
    S_{\mathrm{human}}(A)
      &=\frac{S_{\mathrm{exp}}(A)+S_{\mathrm{vis}}(A)}{2}.
  \end{aligned}
  \label{eq:gamegen-auto-human-scores}
\end{equation}
Across the 15 models, the two aggregates exhibit a strong positive linear
association (Pearson's $r=0.898$, $p=5.38\times10^{-6}$) and a strong rank
association (Spearman's $\rho=0.832$, $p=1.20\times10^{-4}$). As shown in
\Cref{fig:gamegen-human-machine-correlation}, ordinary least-squares regression
yields $S_{\mathrm{human}}=0.643S_{\mathrm{auto}}+25.082$ with
$R^2=0.807$. Thus, agents that realize more required and bonus functionality
generally also deliver games with better player experience and visual quality.
The slope below one further indicates that human-perceived scores vary less
across models than automated functional scores. Nevertheless, the two signals
are not interchangeable: Claude-Opus-4.8 lies above the fitted trend, whereas
DeepSeek-V4-Flash lies below it. These deviations reinforce the need for our
hybrid protocol, as functional coverage and human-perceived quality capture
complementary aspects of generated games rather than duplicate judgments of
the same criteria.

%% file: sections/5_gamefix.tex
\providecommand{\GameFix}{\textsc{GameFix}}
\providecommand{\GameXpert}{\textsc{GameXpert-Bench}}
\providecommand{\strict}{\textsc{Strict}}
\providecommand{\general}{\textsc{General}}
\providecommand{\explicit}{\emph{Explicit Issue}}
\providecommand{\selfdisc}{\emph{Self-Discovery}}
\providecommand{\eg}{e.g.,}
\providecommand{\ie}{i.e.,}

\section{\GameFix: Game Bug Diagnosis and Repair}
\label{sec:gamefix}

\GameFix{} is the bug-repair track of \GameXpert{}. Following the SWE-bench paradigm~\citep{jimenez2024swebench}, we give an agent a real HTML5/JavaScript web game broken by automatically injected bugs and ask it to diagnose and repair the source in a network-disabled sandbox. Each task is graded by a deterministic probe that launches the repaired game in headless Chromium through Playwright and executes behavioral tests. Unlike rubric-based coding benchmarks, this directly verifies game behavior, making the benchmark reproducible, verifiable, inexpensive to extend, and free of LLM-judge bias. We describe the task setting and Gold Games (\S\ref{sec:gf-task}), bug construction (\S\ref{sec:gf-bugs}), evaluation protocol and metrics (\S\ref{sec:gf-eval}), and results on 17 models (\S\ref{sec:gf-results}). Additional examples, test definitions, and complete results are in Appendix~\ref{app:gamefix}.

\subsection{Task setting and gold games}
\label{sec:gf-task}

\GameFix{} is a controlled benchmark built from a curated collection of \textbf{Gold Games}: complete, human-verified, fully playable web games. Bugs are injected through a reversible mutation pipeline (\S\ref{sec:gf-bugs}). Each defect is produced by a mechanically invertible edit whose inverse is the exact gold patch, giving every task fixed ground truth without human annotation and enabling deterministic executable grading.

The closed set contains \textbf{50 independently playable, separately reviewed internal game levels} with distinctive gameplay and control mechanics unavailable on the internet. Each level receives 19--27 bugs at once, so repair requires localisation, prioritisation, and regression avoidance across multiple independent defects within a single session. Combined with the two query modes in \S\ref{sec:gf-eval}, this yields \textbf{100 evaluation tasks} per run. Because the pipeline requires only a gold game and its executable tests, it can be extended to additional games without re-annotation.

\paragraph{Provenance and anti-contamination.} Proprietary games with novel mechanics reduce the risk of dataset contamination. Classic and popular titles are common in pre-training corpora, allowing frontier models to potentially ``repair'' them by recalling memorised source rather than reasoning from observed behavior. In a pilot of \textbf{50 open-source games}, we observed signs of such contamination. Repository quality was also inconsistent, making even ``correct'' behavior difficult to define reliably. Our Gold Games instead use gameplay largely unavailable elsewhere and high player-skill difficulty, making memorisation shortcuts unlikely and requiring the model to reason about playability. Each game is admitted only after \textbf{more than 24 hours of review} and a \textbf{double sign-off} from a game-design specialist and an AI researcher. For confidentiality, the closed set remains internal.

\subsection{Bug construction}
\label{sec:gf-bugs}

Each Gold Game is broken at code sites drawn from a taxonomy of \textbf{7 dimensions and 61 subcategories}, distilled by our game-development team from production experience. The dimensions cover the major ways a game can fail:

\begin{itemize}[leftmargin=*,itemsep=2pt,topsep=2pt]
\item \textit{Core Gameplay}: core mechanics, including controls, combat, AI, physics, and win/lose rules;
\item \textit{Meta}: long-term systems, including progression, economy, achievements, and multiplayer;
\item \textit{UI Design}: HUD, menus, text, layout, tutorials, and accessibility;
\item \textit{Art Design}: visual and audio presentation, including animation, effects, camera, and sound;
\item \textit{Test}: runtime issues, including crashes, freezes, loading failures, and performance;
\item \textit{Level Design}: maps, object placement, level progression, and procedural generation;
\item \textit{Balance Design}: numerical balancing of combat, difficulty, rewards, and economy.
\end{itemize}

Within a chosen subcategory, a bug is created by a small, mechanically reversible mutation operator whose inverse is the gold fix. The pipeline therefore emits both a \texttt{mutation.patch} and its exact inverse, \texttt{gold.patch}. Subcategories specify what kind of behavior breaks, while operators are the low-level edits that cause the breakage. Reversibility provides exact ground truth and makes repair outcomes unambiguous. Worked mutation$\leftrightarrow$gold examples are given in Appendix~\ref{app:gamefix}.

\paragraph{Why many bugs per task.}
Real games rarely present 19--27 defects at once; this count is a deliberate evaluation design rather than a claim about realism. Its purpose is to evaluate long-horizon agentic behavior. In one session, the agent repeatedly performs Diagnose$\rightarrow$Edit$\rightarrow$Test, and we measure how many independent bug sites it can localise, repair, and verify before stopping. This coverage reflects both its understanding of the game and its ability to resolve multiple defects efficiently. Individual bugs are typically one- or two-line edits whose effects often appear only during gameplay. In \selfdisc{}, most symptom descriptions are withheld, further requiring the model to discover bug sites autonomously. Together, these settings evaluate the six agentic abilities defined in Appendix~\ref{app:gf-abilities}.

\subsection{Evaluation protocol}
\label{sec:gf-eval}

\paragraph{Two query modes.} Each task is evaluated under two settings that differ only in how much the prompt discloses:
\begin{itemize}[leftmargin=*,itemsep=2pt,topsep=2pt]
  \item \explicit{} (``list''): the user provides a numbered checklist of all unreasonable behaviours and asks the agent to repair them. The task is to localise and fix each named bug.
  \item \selfdisc{} (``minimal''): the prompt reveals only subjective, presentation-level or taste-dependent symptoms (\eg \textit{art\_design} choices). Objective defects, such as reversed controls, impassable obstacles, or unplayable movement speeds, are withheld. The agent is told that additional strange behaviours remain and must discover and repair them autonomously. The scoring denominator remains the full bug set, so every undiscovered objective bug counts as a failure.
\end{itemize}
The gap between the two modes directly measures a model's \textbf{self-discovery} ability and is a central axis of our analysis.

\paragraph{Executable grading: F2P and P2P.} Each candidate patch is graded by a deterministic program that executes the repaired game. In Node.js, Playwright launches headless Chromium and serves the game locally. A virtual clock advances time in fixed steps, synthetic inputs are dispatched at stage coordinates, and a read-only JSON state snapshot is collected after each step.

Each test case is an executable behavioral check over this state: the evaluator resets the level, advances the clock, dispatches inputs, and evaluates a boolean condition. Parameterized tests must pass for every value in their parameter set. Following SWE-bench~\citep{jimenez2024swebench}, tests are divided into \textbf{Fail-to-Pass (F2P)}, which capture behaviors broken by injected bugs and must pass after repair, and \textbf{Pass-to-Pass (P2P)}, which ensure previously correct behavior does not regress. Each task is evaluated in a single probe session, with every test run from a fresh reset. Tests are grouped by bug site, and the task score is the percentage of bug sites repaired.

\paragraph{Scoring.} A bug counts as \emph{fixed} only when its Fail-to-Pass assertion passes and every associated Pass-to-Pass assertion still passes, ensuring the intended behaviour is restored without regression. Because each task contains 19--27 bugs, we verify at authoring time that no injected bug masks another bug's F2P or breaks its P2P (Appendix~\ref{app:gf-iso}). A P2P failure at grading time therefore reflects a regression introduced by the candidate patch. A task's score is the fraction of bugs fixed,
\begin{equation}
\text{score} \;=\; 100\cdot\frac{\#\text{fixed bugs}}{\#\text{bugs}} \;\in\; [0,100].
\end{equation}

\begin{figure}[t]
  \centering
  \includegraphics[width=\linewidth]{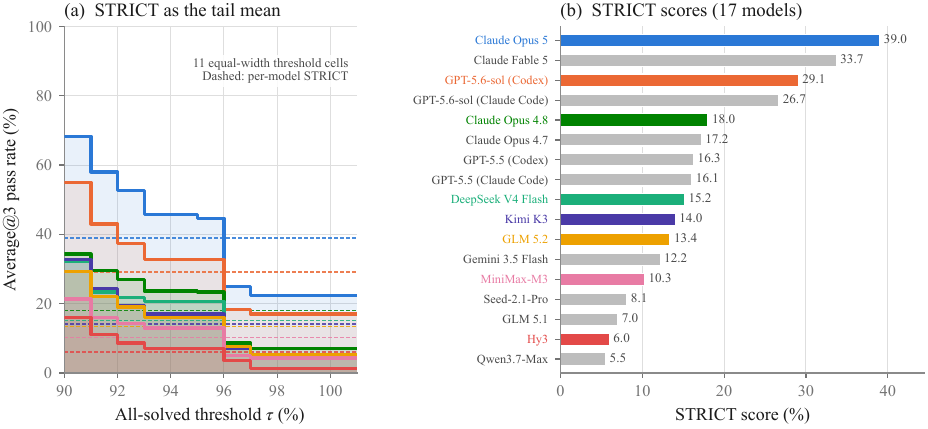}
  \caption{\textbf{Geometric meaning of \strict{}.} Left: for the highlighted
  models, the survival curve over the $\tau\in[90,100]$ band; \strict{} is the
  mean height of that curve across the band (dashed line), equivalently the
  shaded area divided by the band width. Right: \strict{} for
  all 17 models, sorted descending.}
  \label{fig:gf-tailauc}
\end{figure}

\paragraph{From scores to metrics.} For each run $r$ we form a \emph{survival curve} $S_r(\tau)$: for an integer threshold $\tau\in[0,100]$, $S_r(\tau)$ is the percentage of the 100 tasks whose score is at least $\tau$. We report the average of the three curves,
$S(\tau)=\tfrac{1}{3}\sum_r S_r(\tau)$ (Fig.~\ref{fig:gf-tailauc}). Each metric is the mean height of $S$ over a threshold band $[a,b]\subseteq[0,100]$,
\begin{equation}\label{eq:gf-band}
\bar S[a,b] \;=\; \frac{1}{b-a+1}\sum_{\tau=a}^{b} S(\tau),
\qquad a,b\in\mathbb{N}.
\end{equation}

Our primary metric, \strict{}, is the mean height over the near-perfect band $\tau\in[90,100]$,
\begin{equation}
\strict \;=\; \bar S[90,100] \;=\; \frac{1}{11}\sum_{\tau=90}^{100} S(\tau).
\end{equation}
\strict{} is high only when a model repairs nearly every bug in a task across runs, making it sensitive to self-discovery and difficult to saturate.

Alongside \strict{} we report the \textbf{Cliff}, our measure of the
self-discovery deficit. Taking a model's macro average@3 repair score under each
of the two query modes, the Cliff is the drop from \explicit{} to \selfdisc{}.
We report only this difference, not the two absolute scores: under \explicit{}
every bug is named, so absolute scores there are compressed near the top of the
range and separate models poorly, whereas the drop to \selfdisc{} isolates
exactly the ability we care about. All models run with each vendor's maximum reasoning effort in a network-disabled sandbox under the Claude Code agent framework; the two GPT models are additionally run under Codex to separate framework effects from model behavior.

\subsection{Experimental results and analysis}
\label{sec:gf-results}

Table~\ref{tab:gf-leaderboard} gives the full 17-model leaderboard. We draw four
observations from it.

\begin{table}[t]
  \centering
    \caption{\textbf{\GameFix{} leaderboard}, sorted by the primary \strict{}
  metric (mean height of the average@3 survival curve over the band
  $\tau\in[90,100]$).
  Cliff is the drop in macro average@3 repair score from \explicit{} to
  \selfdisc{} (we report this difference only, not the two absolute scores;
  see \S\ref{sec:gf-eval}). All values are 3-run average@3. In each score column \textbf{bold} marks the best value and
  \underline{underline} the worst (Cliff is lower-is-better, so bold is the
  smallest cliff and underline the largest). $^{\dagger}$ models are analysed at the trajectory level in
  Appendix~\ref{app:gamefix}.}
  \label{tab:gf-leaderboard}
  \small
  \setlength{\tabcolsep}{4pt}
  \renewcommand{\arraystretch}{1.05}
  \begin{tabular*}{\linewidth}{@{\extracolsep{\fill}} r l l l r r @{}}
    \toprule
    \# & Model & Harness & Effort & \strict{}$\uparrow$ & Cliff$\downarrow$ \\
    \midrule
    1  & Claude Opus 5$^{\dagger}$        & Claude Code & max   & \textbf{39.0} & \textbf{7.6} \\
    2  & Claude Fable 5                   & Claude Code & max   & 33.7 & 8.4 \\
    3  & GPT-5.6-sol$^{\dagger}$          & Codex & xhigh & 29.1 & 11.3 \\
    4  & GPT-5.6-sol                      & Claude Code & xhigh & 26.7 & 10.3 \\
    5  & Claude Opus 4.8$^{\dagger}$      & Claude Code & max   & 18.0 & 17.9 \\
    6  & Claude Opus 4.7                  & Claude Code & max   & 17.2 & 22.2 \\
    7  & GPT-5.5                          & Codex & xhigh & 16.3 & 13.3 \\
    8  & GPT-5.5                          & Claude Code & xhigh & 16.1 & 10.5 \\
    9  & DeepSeek V4 Flash$^{\dagger}$    & Claude Code & max   & 15.2 & 13.6 \\
    10 & Kimi K3$^{\dagger}$             & Claude Code & max   & 14.0 & 19.6 \\
    11 & GLM 5.2$^{\dagger}$              & Claude Code & xhigh & 13.4 & 20.2 \\
    12 & Gemini 3.5 Flash                 & Claude Code & high  & 12.2 & 31.1 \\
    13 & MiniMax-M3                       & Claude Code & on    & 10.3 & 30.5 \\
    14 & Seed-2.1-pro              & Claude Code & high  & 8.1  & 15.3 \\
    15 & GLM 5.1                          & Claude Code & xhigh & 7.0  & 21.7 \\
    16 & Hy3$^{\dagger}$           & Claude Code & high  & 6.0  & \underline{32.8} \\
    17 & Qwen3.7-Max                      & Claude Code & max   & \underline{5.5}  & 30.3 \\
    \bottomrule
  \end{tabular*}
\end{table}

\begin{figure}[t]
  \centering
  \includegraphics[width=0.60\linewidth]{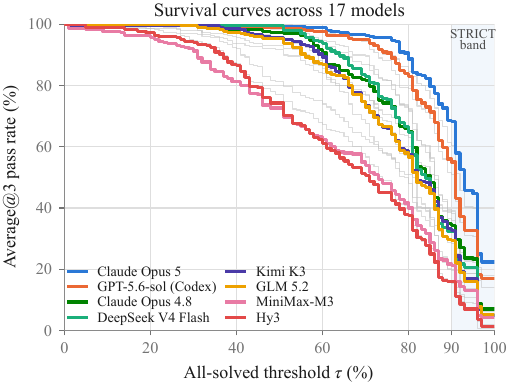}
  \caption{\textbf{Survival curves} of the average@3 pass rate versus the
  all-solved threshold $\tau$ for all 17 models (eight highlighted); each run
  contributes its own curve and the three are averaged. The shaded band
  $\tau\in[90,100]$ is the region whose mean curve height defines \strict{}.}
  \label{fig:gf-survival}
\end{figure}

\begin{figure}[t]
  \centering
  \includegraphics[width=\linewidth]{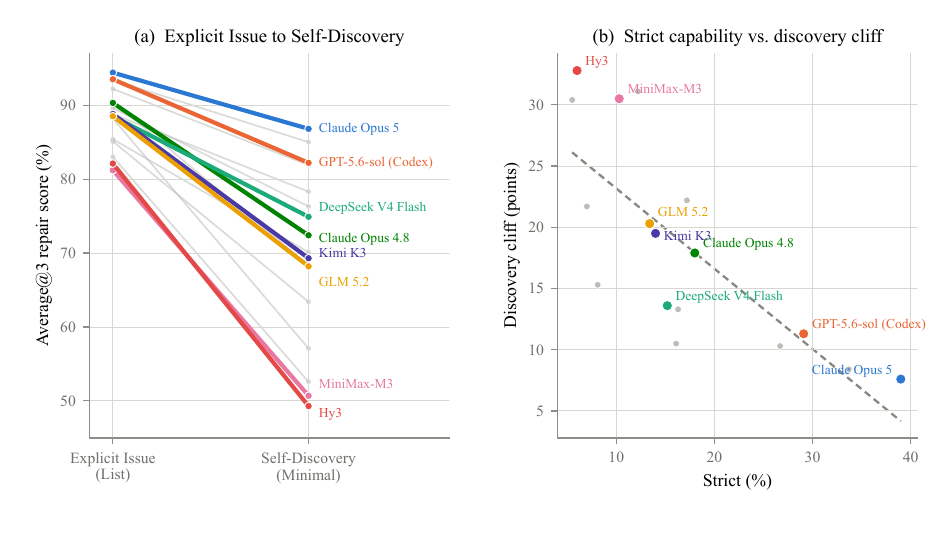}
  \caption{\textbf{The \explicit{}$\rightarrow$\selfdisc{} Cliff.} Left: each
  model's macro average@3 under \explicit{} versus \selfdisc{}; a steeper drop is
  weaker self-discovery. Right (inset): the Cliff shrinks as \strict{} rises.}
  \label{fig:gf-cliff}
\end{figure}

\paragraph{The benchmark is far from saturated.} Even the strongest model, Claude Opus 5, reaches only \strict{}~$=39.0$ out of 100, while the median model sits near 14. Near-perfect multi-bug repair remains rare, leaving substantial headroom.

\paragraph{Models separate under \selfdisc{}, not \explicit{}.} With the full checklist the 17 models are compressed into a span of about 13 points, suggesting that localise-and-fix is close to solved \emph{when every bug is named}. Once the bugs must be discovered autonomously that span widens to about 38 points, nearly three times wider. \strict{} reflects this distinction directly, since reaching the $\ge90$ tail requires recovering nearly all un-hinted bugs (Fig.~\ref{fig:gf-survival}).

\paragraph{The \explicit{}$\rightarrow$\selfdisc{} Cliff narrows as models get stronger.} The Cliff ranges from 7.6 for Opus 5 to 32.8 for Hy-3 and decreases as \strict{} rises (Fig.~\ref{fig:gf-cliff}). A large Cliff indicates that a model can repair bugs once pointed to them but struggles to discover and resolve them autonomously.

\paragraph{Six agentic abilities behind the gap.} The multi-bug setting requires more than code editing. We identify six interacting abilities: \emph{self-discovery}, \emph{behavioural verification}, \emph{value recovery}, \emph{multi-bug coverage and planning}, \emph{regression control}, and \emph{stopping criterion}. Most become substantially more demanding under \selfdisc{}, where the checklist no longer supplies bug locations or a clear completeness target. Appendix~\ref{app:gf-abilities} defines these axes and provides trajectory-level comparisons across representative models.

\paragraph{Two primary mechanisms produce the Cliff.} \emph{(i) Discovery deficit.} Weaker models fail to find many un-hinted bugs, causing their \selfdisc{} score to fall sharply; for example, Hy-3 drops from 96.2 to 34.6 on one instance when the checklist is reduced. \emph{(ii) Fix authorization.} Some models identify additional bugs but decline to repair them because they interpret the checklist as a scope boundary. Claude Opus 4.8, for example, rationalises injected defects as intentional design (100$\rightarrow$59.3 despite similar localisation), while GLM 5.2 identifies a dead enemy subsystem but leaves it as beyond ``minimal changes'' (100$\rightarrow$26.9). In these cases, ``not listed'' is effectively treated as ``not authorised.''

\paragraph{Verification and stopping further limit \strict{}.} Discovery alone is insufficient: models must also infer uncertain values, verify the resulting behaviour, and decide when the task is complete. The sandbox provides no gold reference, so free constants such as physics parameters must often be reconstructed from gameplay constraints or local evidence. Weaker models more often rely on static inspection or plausibility-based checks, while stronger models derive values from in-game invariants and validate them through execution.

\paragraph{Other recurring failure modes.} We also observe \emph{rabbit-holing}, where an agent spends excessive effort on one confusing artefact while leaving other bugs untouched, and occasional \emph{multi-agent conflict}, where parallel workers overwrite or misinterpret one another's edits. These cases are less frequent, but show that long-horizon repair introduces planning and orchestration failures beyond individual bug localisation.

\paragraph{Why stronger models keep the Cliff small.} Stronger models continue searching beyond the disclosed hints, pursue coverage until they have an objective completeness signal, recover missing values from game invariants, and verify repairs against actual execution. Weaker models perform this end-to-end audit less consistently. The resulting reliability gap across discovery, verification, planning, regression control, and stopping is what \strict{} is designed to expose. More trajectory analysis and worked repair cases are provided in Appendix~\ref{app:gamefix}.

%% file: sections/6_gameopt.tex
\section{\GameOpt: Human-Guided Game Optimization}
\label{sec:gameopt}

Game development rarely ends once a game becomes executable. A high-quality game also requires multiple rounds of optimization to enhance its mechanics, level layout, numerical balance, visual language, interface, and audio feedback. 
Unlike bug fixing, these changes generally admit multiple valid implementations and cannot be specified by a single reference patch. 
Therefore, a benchmark for evaluating multi-turn, open-ended game development.

We introduce \GameOpt{}, which is the third stage of our benchmark suite, following game
generation and game bug fixing. It asks a single question: given a real game
and a trajectory-grounded sequence of product requests, can a coding agent
turn a working prototype into a better product without breaking what already
worked?  The games and starting trajectories originate from historical
human--agent co-creation.  Requests retain human requirements where suitable
history exists and use snapshot-grounded synthesis to complete missing design
dimensions; every model receives the same fixed replay inputs.

\subsection{Task Formulation}
\label{sec:gameopt-task}

A \GameOpt{} instance consists of an initial game repository $G^{(0)}$ and an
ordered sequence of natural-language requests
$U=(u_1,\ldots,u_T)$.  At turn $t$, the coding agent receives the current
repository and the new request, and produces a set of edits $\Delta_t$:
\begin{equation}
    \Delta_t = \mathcal{A}\!\left(G^{(t-1)},u_t,h_{<t}\right),
    \qquad
    G^{(t)} = \operatorname{Apply}\!\left(G^{(t-1)},\Delta_t\right)
    \label{eq:gameopt-transition}
\end{equation}
The repository is never reset between turns, so the chain is a genuine
long-horizon process rather than six independent edit tasks: the state on
which request $u_{t}$ is served is whatever the agent itself produced in the
first $t-1$ turns.

Each turn targets one of six design dimensions,
\begin{equation}
  \mathcal{D}=\{\textsc{gameplay},\,\textsc{level},\,\textsc{balance},\,
                \textsc{art},\,\textsc{ui},\,\textsc{audio}\},
  \qquad |\mathcal{D}|=T,
\end{equation}

and a chain-specific bijection $\pi_c:\{1,\dots,T\}\!\to\!\mathcal{D}$ assigns
one dimension to each turn. Conceptually, the agent is asked to solve
\begin{equation}
    \max_{G^{(T)}}
    \sum_{d\in\mathcal{D}} \alpha_d Q_d\!\left(G^{(T)}\right)
    - \lambda\,\mathcal{R}\!\left(G^{(0)},G^{(T)}\right)
    \label{eq:gameopt-objective}
\end{equation}
where $Q_d$ is not assumed to be a unique ground-truth quality function;
different implementations may realize the same user intent.  In the
benchmark, it is operationalized by task-specific acceptance criteria and
observable evidence.  The regression term $\mathcal{R}$ captures the equally
important requirement that the final game remain buildable, playable, and
compatible with its original core loop.

The target use case is live human--agent collaboration, in which a person
plays or inspects successive versions and expresses the next need from a
player or product perspective.  For controlled model comparison, however,
\GameOpt{} instantiates this setting as an offline replay: all models receive
the same ordered requests and the same starting snapshot.  This design
retains the cumulative nature of human-guided optimization while removing
variation caused by different users choosing different follow-up requests.

\subsection{Trajectory-Based Benchmark Construction}
\label{sec:gameopt-data}

\paragraph{Source trajectories.}
The data originate from historical game co-creation trajectories between
users and coding agents.  Each trajectory records a sequence of user
requests together with the code artifact produced after each request.  We
select a playable intermediate version as the starting snapshot: it must be
sufficiently complete to admit meaningful optimization, while the requested
improvements must still be observable or implementable from that snapshot.
Only this selected snapshot is exposed during evaluation; later historical
versions are not provided to the tested model.

Each benchmark chain contains six turns, one for every dimension in
Eq.~\eqref{eq:gameopt-objective}.  When a historical trajectory contains a
suitable request, the user request is retained and lightly normalized for
clarity.  Dimensions absent from the trajectory are completed with
strong-model-generated requests grounded in concrete deficiencies of the
same snapshot.  Synthetic requests are therefore used for coverage rather
than as generic, repository-independent instructions.  The resulting task
prompt preserves intentional product-level ambiguity (e.g., ``make the
encounter more tense''), whereas a normalized intent and the acceptance
criteria are kept private from the tested model.

\paragraph{Dataset composition.}
The current \GameOpt{} evaluation collection contains 17 self-contained
JavaScript games, each with one six-turn chain, for 102 optimization turns in
total.  Its 701 acceptance criteria comprise 604 positive rubric items and 97
regression checks.  Table~\ref{tab:gameopt-composition} reports the composition;
the complete chain inventory appears in Appendix~\ref{app:gameopt-overview}.

\begin{table}[t]
  \centering
  \small
  \setlength{\tabcolsep}{7pt}
  \renewcommand{\arraystretch}{1.18}
  \caption{Composition of the current \GameOpt{} evaluation collection.}
  \label{tab:gameopt-composition}

  \begin{tabular}{@{}lr@{}}
    \toprule
    & \textbf{Count} \\
    \midrule
    Games (chains)      &  17 \\
    Turns               & 102 \\
    Acceptance criteria & 701 \\
    \addlinespace[2pt]
    \quad Requirement   & 392 \\
    \quad Challenge     & 212 \\
    \quad Regression    &  97 \\
    \bottomrule
  \end{tabular}
\end{table}

\paragraph{Evidence-oriented rubrics.}
For each turn, the benchmark defines a hidden set of independently testable
rubric items.  A positive item is either a \emph{requirement}, which captures
the user's stated intent, or a \emph{challenge}, which checks integration and
edge cases that are necessary for a robust implementation.  A
\emph{regression} item instead detects functionality present in $G^{(0)}$
that is broken in $G^{(T)}$.  Each item specifies its priority, provenance,
evaluation modality (code, rendered output, or both), and the evidence
required for a judgment.  Compound conditions follow an
all-conditions-required rule: an item passes only when every explicit
condition is established.  Comments, unused configuration, dead code, and
the model's own description of its changes are not accepted as evidence.

The current bundle contains 604 positive items (392 requirements and 212
challenges) and 97 regression checks.  Of all 701 items, 281 are judged from
code, 122 require both code and rendered evidence, and 298 use rendered
evidence.  Thirty-three items are explicitly marked as proxies for properties such as
visual coherence or game feel.  This marker is important: the presence of an
implementation path can support a proxy criterion, but does not by itself
prove that the resulting experience is aesthetically superior.

\subsection{Evaluation Protocol and Scoring}
\label{sec:gameopt-protocol}
\paragraph{Replay.}
For every model and chain, the harness copies $G^{(0)}$ into an isolated
workspace, records file hashes, and runs a preflight check whose findings are
logged as pre-existing conditions and are never attributed to the model. At
turn $t$ only the prompt $u_{t}$ is released; the model continues from its own
$G^{(t-1)}$ with no access to the criteria, to a reference implementation, or
to any later human version. After the sixth turn the workspace is frozen and
the complete diff against $G^{(0)}$ is recorded.

\paragraph{Final-product-only judging.}
All six turns are evaluated on the frozen $G^{(T)}$.  A positive criterion $i$
is binary, with no partial credit, and passes only when every explicit
condition in its description is established by admissible evidence:
\begin{equation}
  x_{m,k,i}=\prod_{j\in\mathcal{C}_i}
      \mathbf{1}\!\left[\text{condition } j \text{ holds in } G^{(T)}\right]
      \cdot
      \mathbf{1}\!\left[\text{evidence for } j \text{ is admissible}\right].
  \label{eq:gameopt-item}
\end{equation}
Here $x_{m,k,i}\in\{0,1\}$ is the judgment for model $m$, full-corpus round
$k$, and positive item $i$.  Admissible evidence is a code location on a
reachable call path, a runtime log, a screenshot, or an audio trace; comments,
unused configuration, dead code, and the model's own description are not
admissible, and insufficient evidence resolves to $x_{m,k,i}=0$.

\paragraph{Difficulty-leaning discrimination weights.}
Let $K=3$ and $\mathcal{P}$ be the 604 positive items pooled across all 17
chains.  For item $i$, let $n_i$ be its number of judgments and
$n_i^{\mathrm{pass}}$ its number of passes.  We assign a common weight
\begin{equation}
  p_i=\frac{n_i^{\mathrm{pass}}}{n_i},
  \qquad
  w_i=p_i(1-p_i)^2.
  \label{eq:gameopt-weight}
\end{equation}
The $p_i(1-p_i)$ factor rewards items that discriminate among systems, while
the additional $(1-p_i)$ factor tilts weight toward harder items.  Items passed
by everyone or no one receive zero weight.  A model's positive-item score in
round $k$ is the pooled weighted pass rate
\begin{equation}
  S^{+}_{m,k}=100\,
  \frac{\sum_{i\in\mathcal{P}}w_i x_{m,k,i}}
       {\sum_{i\in\mathcal{P}}w_i}.
  \label{eq:gameopt-round-score}
\end{equation}
Pooling is performed over the full item bank rather than by averaging a score
for each game first; consequently, chains containing more discriminative hard
items contribute more weight.  Let $\mathcal{R}$ be the applicable regression
rows, $r_{m,k,j}\in\{0,1\}$ indicate whether regression $j$ is triggered, and
$q_j<0$ be its predefined penalty.  The reported score is
\begin{equation}
  S_{m,k}=S^{+}_{m,k}
  +\sum_{j\in\mathcal{R}}q_j r_{m,k,j}.
  \label{eq:gameopt-regression-score}
\end{equation}
Thus, requirement and challenge items contribute positive credit, while
triggered regression rows deduct points.  Priority-specific, dimensional, and
turn-specific positive-item values use the same weighted-pass formula after
restricting $\mathcal{P}$ to the corresponding subset.  There is no P0 gate or
runtime cap in the reported leaderboard.  Build/start failures, dead products,
and core-loop failures are also reported as separate diagnostics.

\paragraph{Result aggregation.}
The reported overall, dimension-specific, and turn-specific scores use the
same unified result summary for every model.  We retain the three round values
to expose run-to-run variation; chain-level statistical comparisons should
treat the chain, rather than an individual rubric item, as the independent
unit.

\subsection{Experimental Results and Analysis}
\label{sec:gameopt-results}

\paragraph{Setup.}
We report 15 model variants on the 17-game JavaScript collection, with six
turns per game and three full-corpus evaluation rounds.  This yields 102 turns
per model and 765 model--game runs in total.  Overall, dimension-specific, and
turn-specific results use the same 15-model cohort.

\begin{table}[H]
  \centering
  \footnotesize
  \setlength{\tabcolsep}{3.4pt}
  \renewcommand{\arraystretch}{1.06}
  \caption{Overall and six-dimensional \GameOpt{} scores for 15 model variants
  on 17 JavaScript games.  All columns use the pooled
  discrimination-weighted result summary.}
  \label{tab:gameopt-main}
  \begin{tabular}{@{}rlrrrrrrr@{}}
    \toprule
    \textbf{Rank}
    & \textbf{Model}
    & \textbf{Overall}
    & \textbf{Gameplay}
    & \textbf{Level}
    & \textbf{Balance}
    & \textbf{Art}
    & \textbf{UI}
    & \textbf{Audio} \\
    \midrule
     1 & Claude-Opus-5                 & {93.96} & 91.3 & 94.5 & 80.8 & 100.0 & 97.0 & 98.5 \\
     2 & Claude-Fable-5                & 89.31 & 88.7 & 85.2 & 80.9 & 95.8 & 90.3 & 96.2 \\
     3 & Kimi-K3                       & 84.66 & 86.1 & 75.9 & 81.1 & 91.6 & 83.6 & 93.9 \\
     4 & GPT-5.6-sol                   & 83.68 & 82.8 & 89.7 & 83.2 & 87.7 & 67.8 & 97.0 \\
     5 & Claude-Opus-4.8               & 82.53 & 75.3 & 78.3 & 77.8 & 77.9 & 95.1 & 93.5 \\
     6 & Claude-Opus-4.7               & 82.36 & 87.7 & 83.6 & 89.4 & 79.6 & 66.8 & 94.8 \\
     7 & DeepSeek-V4-Flash             & 78.34 & 71.9 & 67.3 & 70.2 & 87.2 & 90.5 & 83.9 \\
     8 & GPT-5.5                       & 74.24 & 70.5 & 67.0 & 86.4 & 66.5 & 78.7 & 83.7 \\
     9 & GLM-5.2                       & 72.98 & 74.5 & 68.3 & 63.1 & 67.8 & 79.1 & 88.3 \\
    10 & Hy3                     & 68.87 & 71.1 & 62.0 & 58.8 & 66.4 & 75.5 & 82.1 \\
    11 & Qwen3.7-Max                   & 64.77 & 67.7 & 55.6 & 54.4 & 65.1 & 72.0 & 76.0 \\
    12 & GLM-5.1                       & 60.67 & 64.3 & 49.2 & 50.1 & 63.7 & 68.4 & 69.8 \\
    13 & MiniMax-M3                    & 59.87 & 71.4 & 49.8 & 53.9 & 48.8 & 70.2 & 64.7 \\
    14 & Gemini-3.5-Flash              & 47.20 & 52.7 & 36.6 & 33.1 & 53.9 & 50.4 & 58.2 \\
    15 & Seed-2.1-Pro                  & 35.89 & 46.8 & 32.6 & 28.6 & 28.7 & 32.8 & 49.0 \\
    \bottomrule
  \end{tabular}
\end{table}

\paragraph{Overall results.}
Claude-Opus-5 ranks first at 93.96, followed by Claude-Fable-5 at 89.31 and
Kimi-K3 at 84.66.  Across all 15 models, the median is 74.24,
and the 58.07-point
range from Claude-Opus-5 to Seed-2.1-Pro shows that the difficulty-leaning
weighting separates the field substantially.  Round stability also varies
sharply.  GPT-5.6-sol spans only 0.13
points across rounds, whereas Gemini-3.5-Flash and MiniMax-M3 span 36.12 and
33.67 points.  The six-dimensional columns additionally expose whether similar
overall scores arise from balanced capability or from sharply different
strengths across gameplay, level design, balance, art, UI, and audio.

\begin{figure}[t]
  \centering
  \includegraphics[width=\textwidth]{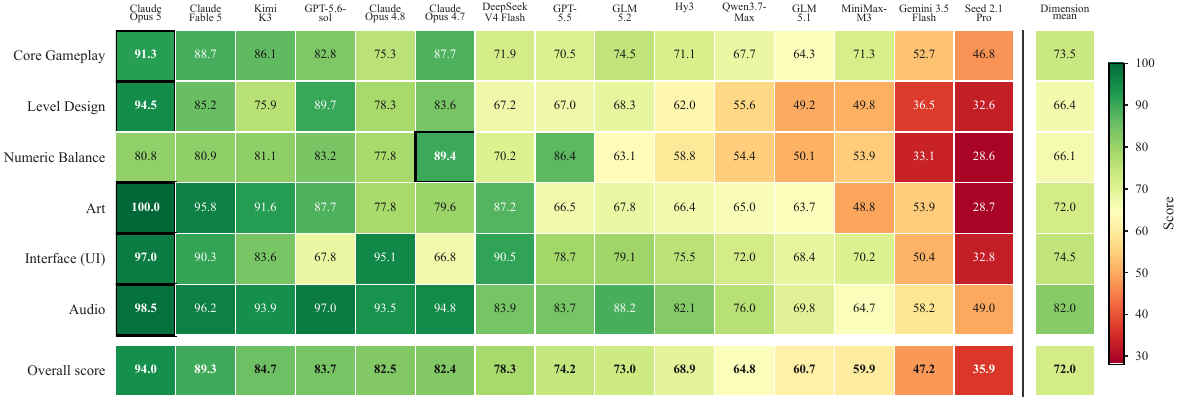}
  \caption{\textbf{Six-dimensional capability profiles.} Rows denote the six
  optimization dimensions and columns denote models.  Each cell reports the
  final-product rubric score; black outlines mark the best-performing model in
  each dimension.  The right column reports means over all 15 models, while the
  bottom row reports the leaderboard Overall score.}
  \label{fig:gameopt-sixdim}
\end{figure}

\paragraph{Six-dimensional capability.}
Figure~\ref{fig:gameopt-sixdim} reveals structure hidden by the overall score.
Across all 15 models, numeric balance (66.11) and level design (66.36) have the
lowest means, whereas audio is highest (81.97).  Numeric balance is the weakest
dimension for 7/15 models and level design for 3/15; audio is the strongest for
10/15.  High overall performance does not imply an
even profile: Claude-Opus-5 ranges from 80.80 in balance to 100.00 in art, and
GPT-5.6-sol spans 29.19 points between UI and audio.  These profiles are
descriptive rather than causal, but they show why the pooled Overall score
should be accompanied by dimension-specific results.

\begin{figure}[h]
  \centering
  \includegraphics[width=\textwidth]{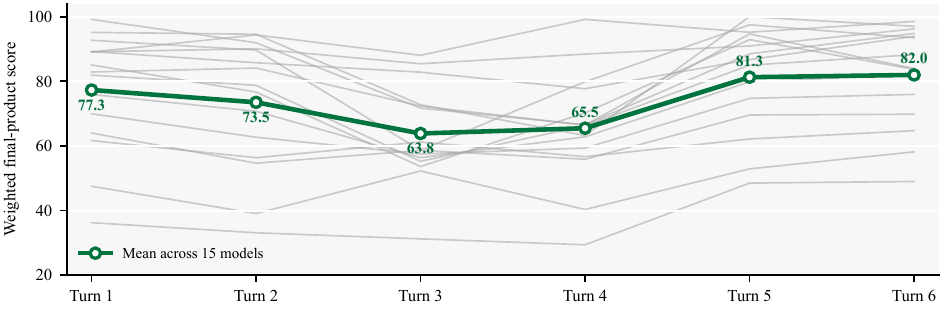}
  \caption{\textbf{Performance across six optimization turns.} Each value is
  the discrimination-weighted final fulfillment score for requests introduced
  at the corresponding turn.  Thin gray curves show individual models and the
  green curve is their macro-average.}
  \label{fig:gameopt-turns}
\end{figure}

\paragraph{Multi-turn behavior.}
The model macro-average follows a pronounced non-monotonic pattern: 77.30,
73.50, 63.85, 65.48, 81.27, and 81.97 from turns one through six.  Every model
reaches its lowest value at turn two, three, or four (3, 7, and 5 models,
respectively), and none is weakest at turn one, five, or six.  This
rules out a simple monotonic turn-wise decay in the current data, but it does
not establish long-context stability.  Turn and content are confounded---for
example, audio is always turn six, while balance is concentrated at turn three
and art at turn four---so the rebound at turns five and six cannot be
attributed to accumulated context.  Longer, counterbalanced trajectories are
needed to isolate retention from request and dimension difficulty.

\begin{figure}[h]
  \centering
  \includegraphics[width=\textwidth]{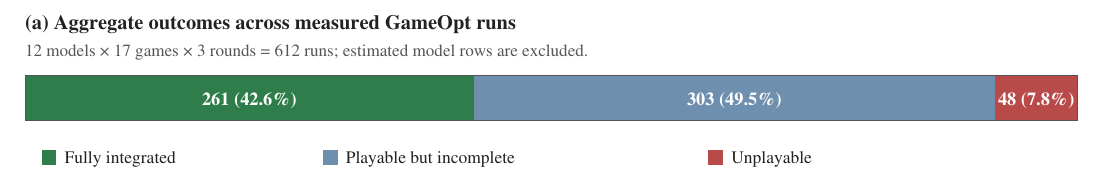}
  \vspace{0.5em}
  \includegraphics[width=\textwidth]{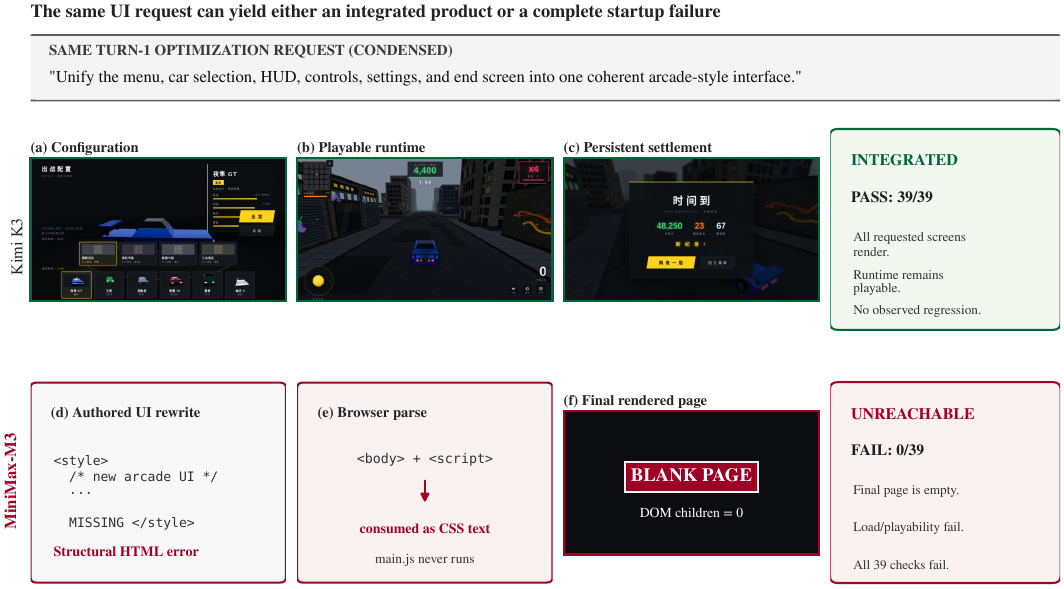}
  \caption{Aggregate integration outcomes and representative
  extremes and case study.}
  \label{fig:gameopt-cases}
\end{figure}

\paragraph{Case study: authored features versus integrated products.}
Figure~\ref{fig:gameopt-cases} places the paired examples in their aggregate
context by separating final products into fully integrated,
playable-but-incomplete, and unplayable outcomes.  The examples then control
the requested change by contrasting two models on the same
first-turn interface optimization.  kimi-K3's final product
makes the redesigned selection screen, runtime HUD, and settlement state all
reachable, demonstrating that the interface change remains connected to the
game loop.  MiniMax-M3 also authors the requested arcade-style CSS, but a
single structural omission---the closing \texttt{</style>} tag---changes how
the browser parses the entire document.  The body has no rendered children,
\texttt{main.js} never executes, and the same blank frame is observed at the
menu, selection, HUD, and settlement probes.  Thus, every requested dimension
becomes unreachable even though the feature code is present.  This paired case
explains the mechanism behind the tails of the aggregate distribution:
optimization quality
depends on an executable, end-to-end product, not merely on locally plausible
edits.

%% file: sections/7_conclusion.tex
\section{Conclusion}
\label{sec:conclusion}

We introduced \GameXpert{}, an execution-grounded benchmark for evaluating
coding agents across game generation, diagnosis and repair, and human-guided
optimization. Across all three stages, the central finding is that producing
an apparently plausible implementation is easier than delivering a rich,
verified, and regression-free game. By evaluating not only what agents write
but also what their games actually do, \GameXpert{} provides a foundation for
developing agents that can participate reliably in the full game-development
lifecycle rather than only produce its first playable draft.

%% file: appendices/1_gamegen_details.tex
\section{Additional Details of GameGen}
\label{app:gamegen}

This appendix provides additional details of the \GameGen{} corpus and its
evaluation. We report the category distribution, representative full prompts,
the complete Shared Rubric for one example game, complete 2D and 3D subset
leaderboards, and per-game event pass rates. All source material originally
written in Chinese is translated into English.

\subsection{Corpus Statistics and Representative Prompts}
\label{app:gamegen-prompts}

\paragraph{Category statistics.}
\GameGen{} contains 97 games from 11 categories, comprising 53 2D games and 44
3D games. As shown in \Cref{tab:gamegen-category-statistics}, most categories
contain both 2D and 3D tasks. Simulation and Management is the only category
without a 3D task in the current corpus.

\begin{table*}[h]
\centering
\caption{Category distribution of the \GameGen{} corpus. The Share column is
computed over all 97 games.}
\label{tab:gamegen-category-statistics}
\small
\setlength{\tabcolsep}{8pt}
\begin{tabular}{lrrrr}
\toprule
\textbf{Category} & \textbf{2D} & \textbf{3D} & \textbf{Total} & \textbf{Share (\%)} \\
\midrule
Puzzle                    & 5 & 5 & 10 & 10.3 \\
Strategy                  & 5 & 5 & 10 & 10.3 \\
RPG                       & 5 & 5 & 10 & 10.3 \\
Shooter                   & 5 & 4 &  9 &  9.3 \\
Roguelike                 & 3 & 3 &  6 &  6.2 \\
Simulation and Management & 5 & 0 &  5 &  5.2 \\
Action                    & 5 & 5 & 10 & 10.3 \\
Narrative                 & 5 & 3 &  8 &  8.2 \\
Idle                      & 5 & 5 & 10 & 10.3 \\
Sandbox and Building      & 5 & 5 & 10 & 10.3 \\
Card                      & 5 & 4 &  9 &  9.3 \\
\midrule
\textbf{Total}            & \textbf{53} & \textbf{44} & \textbf{97} & \textbf{100.0} \\
\bottomrule
\end{tabular}
\end{table*}

\paragraph{Representative full prompts.}
Below, we provide one complete 2D prompt and one complete 3D prompt from each
category. Since Simulation and Management contains no 3D task, only its 2D
prompt is reported. The prompts are translated without omitting any requested
gameplay feature.

\begin{table*}[p]
\centering
\caption{Representative full prompts from \GameGen{} (Part I of III).}
\label{tab:gamegen-representative-prompts-1}
\footnotesize
\renewcommand{\arraystretch}{1.12}
\setlength{\tabcolsep}{4pt}
\begin{tabular}{p{0.14\textwidth}cp{0.72\textwidth}}
\toprule
\textbf{Category} & \textbf{Dim.} & \textbf{Full Prompt} \\
\midrule
Puzzle & 2D & Create a 2048 game controlled with the arrow keys. Tiles with the same value should merge, and a new tile with value 2 or 4 should be spawned at random after every valid move. Include sliding animations during movement and merging, display the current score, show a victory message when the player reaches 2048, and show a game-over message when no valid move remains. \\
\addlinespace
Puzzle & 3D & Create a minimalist, wood-textured 3D maze game viewed from above. The game should support tilting the maze to guide a ball through a wooden map containing walls and an exit. The player tilts the maze to roll the ball, avoids traps and dead ends, and guides the ball to the exit. The level fails if the ball falls out of the maze or time expires. Include at least two basic levels: a single-route maze and a simple branching maze. \\
\addlinespace
Strategy & 2D & Create a hand-painted watercolor-style 2D insect-themed real-time strategy game viewed from above. Include an economy based on collecting honeydew and cultivating fungus, an ant queen that produces workers with different roles, and switching between surface and underground maps. The player controls an ant colony, expands an underground nest and its tunnels, sends worker ants to collect food, and deploys soldier ants against invading spiders or rival ant colonies. \\
\addlinespace
Strategy & 3D & Create a post-apocalyptic science-fiction 3D first-person tower-defense game. Include defensive structures and a weapon-upgrade system. The player builds defenses, upgrades weapon systems, and withstands waves of monsters. Enemies should approach the base along different routes. Include at least two of the following defensive structures: a machine-gun turret, a laser turret, and a freezing turret. \\
\addlinespace
RPG & 2D & Create a retro-JRPG-style 2D pixel-art turn-based role-playing game viewed from above. Include companion recruitment and skill-combination systems, with an open map containing both a town and a dungeon. The player can recruit companions, combine skills, and trigger side quests. NPC attitudes should change according to the player's choices. Include at least three character classes: warrior, mage, and rogue. \\
\addlinespace
RPG & 3D & Create a third-person 3D action RPG in an ink-wash wuxia style that fuses ukiyo-e and Chinese ink-painting aesthetics. The environments should include a bamboo forest, a snowy mountain, and a ruined shrine. The core mechanic is a block-and-parry system: a perfect parry should inflict substantial posture damage, and breaking an enemy's posture should enable a finishing move. Allow the player to switch among three prosthetic tools---shuriken, firecrackers, and an axe---and combine light and heavy attacks into different combos. Enemies should include a samurai general whose AI switches between high and low perilous attacks, and a vengeful spirit surrounded by ghost fire that can spread to and burn the player. \\
\addlinespace
Shooter & 2D & Create a 2D top-down aircraft shooter. The player can move the aircraft to dodge enemy bullet patterns, use multiple firing modes and screen-clearing bombs, and fight enemies. Defeating elite enemies should drop firepower upgrades, bombs, and auxiliary weapons, while defeating a boss should award an extra life. Firepower upgrades, extra lives, auxiliary weapons, and bombs must carry over between stages. Their effects and upgrade rules should follow the conventions of the \emph{Raiden} series. \\
\addlinespace
Shooter & 3D & Create a simple cartoon-style 3D first-person shooter set in a small enclosed indoor map. Include firearms and AI-controlled enemies. The player should be able to move, aim, and fire. Enemies should patrol along fixed routes. Include at least two weapons: a pistol and a rifle. \\
\bottomrule
\end{tabular}
\end{table*}

\begin{table*}[p]
\centering
\caption{Representative full prompts from \GameGen{} (Part II of III).}
\label{tab:gamegen-representative-prompts-2}
\footnotesize
\renewcommand{\arraystretch}{1.12}
\setlength{\tabcolsep}{4pt}
\begin{tabular}{p{0.14\textwidth}cp{0.72\textwidth}}
\toprule
\textbf{Category} & \textbf{Dim.} & \textbf{Full Prompt} \\
\midrule
Roguelike & 2D & Create a pixel-art 2D top-down dungeon-exploration roguelike with randomly generated rooms, permadeath, and relic collection. Each run should generate a new room layout, enemy distribution, and set of treasures. The character is permanently lost upon death, but soul stones can be carried back to upgrade global talents. Place one boss every three floors. Include at least two classes: warrior and mage. \\
\addlinespace
Roguelike & 3D & Create a 3D zombie-survival roguelike set in a ruined city, with randomized weapons, ammunition management, and an infection meter. The player searches through zombie hordes for randomized weapons, including pistols, rifles, and machine guns. After each wave, the player can upgrade a held weapon by increasing reserve ammunition or damage, or heal the character to reduce the infection meter. Zombie bites increase infection; when the meter is full, the player turns into a zombie and the run ends. Include at least two zombie types: normal zombies and sprinting zombies. \\
\addlinespace
Simulation and Management & 2D & Create a farm-management game. Customers should arrive automatically to buy crops, while the player plants and harvests produce and hires workers to help with watering and fertilizing. Customers pay coins after completing a purchase. Coins can be used to upgrade the farm, buy new seeds, and unlock animals. Allow the player to change farm decorations dynamically. Crop hybridization should have a random chance of failure, and hired workers should have a fatigue system. \\
\addlinespace
Action & 2D & Create a pixel-art 2D side-scrolling action game set in a gothic castle. Include a combo system, dodge rolls, and a special attack. The player can combine light and heavy attacks into combos, and spend energy to unleash a full-screen special attack. Dodge rolls should provide invincibility frames that can avoid incoming attacks. Include at least two weapon types: a sword and a whip. \\
\addlinespace
Action & 3D & Create a high-speed 3D character-action hack-and-slash game inspired by \emph{Devil May Cry}. Its core mechanics should include real-time switching among four combat styles, aerial combos, enemy-step jumps that cancel action recovery, and a chargeable demonic state that changes move properties and adds damage. Levels should contain hidden challenge rooms; clearing one within the time limit should unlock an additional temporary attribute enhancement. \\
\addlinespace
Narrative & 2D & Create an urban rule-horror text game. The player must avoid supernatural corruption, inspect blurred and crossed-out clauses on an elevator notice and blood-red writing on office notes, investigate strange shadows in surveillance blind spots, identify false rules mixed among genuine ones, and escape the out-of-control office building safely before midnight. \\
\addlinespace
Narrative & 3D & Create a 3D text-driven treasure-hunting game set in an ancient desert city. The vast three-dimensional desert should feel open and desolate. The player traverses dunes and the Gobi, explores three-dimensional ancient-city ruins, deciphers writing on old stone walls, avoids sandstorm hazards, breaks the city's ancient seals, and searches for a legendary treasure. Include underground-palace mechanism puzzles and sandstorm-survival gameplay. \\
\bottomrule
\end{tabular}
\end{table*}

\begin{table*}[p]
\centering
\caption{Representative full prompts from \GameGen{} (Part III of III).}
\label{tab:gamegen-representative-prompts-3}
\footnotesize
\renewcommand{\arraystretch}{1.12}
\setlength{\tabcolsep}{4pt}
\begin{tabular}{p{0.14\textwidth}cp{0.72\textwidth}}
\toprule
\textbf{Category} & \textbf{Dim.} & \textbf{Full Prompt} \\
\midrule
Idle & 2D & Create a retro pixel-art 2D idle game viewed from above and set in a hero camp. Include offline earnings and a skill tree. The player clicks monsters to deal damage and spends coins to purchase units that attack automatically. The hero can upgrade skills. Include at least two automated unit types: archers and mages. \\
\addlinespace
Idle & 3D & Create a 3D retro-steampunk idle game built around three core mechanics: a mechanical factory that automatically forges devices, steam-powered automata that automatically leave to work, and a power core that automatically stores and accumulates energy. Offline idling should mass-produce industrial items. Mechanical modification should carry a risk of malfunction; the player can allocate repair fuel to maintain stable operation and trade precision blueprints through a steam-powered merchant guild to build advanced mechanical creations. \\
\addlinespace
Sandbox and Building & 2D & Create a pixel-art 2D top-down sandbox-building game with terrain editing, resource gathering, and construction in a procedurally generated infinite world. The player can mine and place multiple block types---soil, stone, wood, and water---and build houses, farms, and defensive structures. Include at least two biomes: grassland and desert. \\
\addlinespace
Sandbox and Building & 3D & Create a low-poly 3D first-person sandbox-building game with block placement and removal, a day--night cycle, and hostile creatures on a procedurally generated continent. The player gathers wood and stone and builds houses and castles. Zombies and skeletons should spawn at night. Include at least two biomes: forest and mountain. \\
\addlinespace
Card & 2D & Create a dark-fantasy 2D pixel-art roguelike card game viewed from above, with a randomized room-based dungeon map and deckbuilding mechanics. The player can collect cards, construct a deck, and trigger combo effects. Enemies should patrol rooms and attack proactively. Include at least three card types: attack, defense, and utility cards. \\
\addlinespace
Card & 3D & Create a 3D medieval knight battle-card game set in retro castle and field environments, with an action-oriented main storyline about a kingdom at war. Allow the player to recruit heavy knights, ranger scouts, and royal mages as combat cards. Include unit-type counters and formation-charge mechanics. The player deploys cards to arrange offensive and defensive formations and engages in card-based duels between lords. Pacifying the conflict should unlock additional chapters of the kingdom campaign. \\
\bottomrule
\end{tabular}
\end{table*}

\subsection{Example Shared Rubric: 2048}
\label{app:gamegen-2048-rubric}

To illustrate the event-level evaluation used by \GameGen{}, we provide the
complete Shared Rubric for the 2048 task. Core events determine Completeness,
whereas bonus events determine Richness. The two components are reported
separately in \Cref{tab:gamegen-2048-completeness,tab:gamegen-2048-richness}.

\begin{table*}[p]
\centering
\caption{Completeness checklist for the 2048 task.}
\label{tab:gamegen-2048-completeness}
\small
\renewcommand{\arraystretch}{1.12}
\setlength{\tabcolsep}{4pt}
\begin{tabular}{cp{0.27\textwidth}p{0.62\textwidth}}
\toprule
\textbf{\#} & \textbf{Checklist Item} & \textbf{Verification Criterion} \\
\midrule
1 & Score increases after a merge & Add the value of the newly merged tile to the score and update the visible score display. \\
\addlinespace
2 & An invalid move leaves the state unchanged & If no tile can move or merge in the requested direction, do not spawn a tile or change the score. \\
\addlinespace
3 & Game over when no move remains & End the game when the board is full and no adjacent equal-valued tiles can be merged. \\
\addlinespace
4 & High-score update and persistence & Update the historical best score and retain it after restarting or refreshing the page. \\
\addlinespace
5 & Game start and initialization & Construct the board, reset the score and state, and spawn the initial tiles when the page loads. \\
\addlinespace
6 & Equal-valued tiles merge & Merge two adjacent tiles with the same value along the movement direction into one tile with twice the value. \\
\addlinespace
7 & A new tile appears after a valid move & After any move that changes the board, spawn a new tile in an empty cell. \\
\addlinespace
8 & Reaching 2048 triggers victory & Display a victory state when a tile with value 2048 is created. \\
\addlinespace
9 & Restarting begins a fresh game & The new-game control clears the board, resets the current score and overlays, and creates a new initial state. \\
\addlinespace
10 & Tiles visibly slide between cells & Use a smooth positional transition during movement rather than an instantaneous jump. \\
\bottomrule
\end{tabular}
\end{table*}

\begin{table*}[p]
\centering
\caption{Richness checklist for the 2048 task.}
\label{tab:gamegen-2048-richness}
\small
\renewcommand{\arraystretch}{1.12}
\setlength{\tabcolsep}{4pt}
\begin{tabular}{cp{0.27\textwidth}p{0.62\textwidth}}
\toprule
\textbf{\#} & \textbf{Checklist Item} & \textbf{Verification Criterion} \\
\midrule
1 & Continue after reaching 2048 & Allow the player to dismiss the victory overlay and continue playing beyond 2048. \\
\addlinespace
2 & Touch-swipe controls & Infer a direction from a sufficiently long touch gesture and execute the corresponding move. \\
\addlinespace
3 & WASD controls & Map the WASD keys to the same four movement directions as the arrow keys. \\
\addlinespace
4 & Spawn and merge tween effects & Animate newly spawned tiles and tiles created by merging with distinct appearance or pop effects. \\
\addlinespace
5 & Styles for tiles above 2048 & Provide a visually distinct high-tier style for values greater than 2048. \\
\addlinespace
6 & Multiple merges in one move & Correctly merge four equal-valued tiles in a row into two doubled tiles during a single move. \\
\addlinespace
7 & Pop-in animation for created tiles & Apply a scale or appearance animation to newly spawned or merged tiles. \\
\addlinespace
8 & Resume state after reloading & Restore the board, score, and relevant game state from persistent browser storage. \\
\addlinespace
9 & Input locking during animation & Ignore additional movement commands while a movement animation is in progress. \\
\addlinespace
10 & Merge-bounce animation & Apply a short scaling or bouncing animation to a tile created by merging. \\
\addlinespace
11 & On-screen directional controls & Provide clickable directional buttons as an alternative input method. \\
\addlinespace
12 & Pointer-drag gesture controls & Detect a pointer drag beyond a distance threshold and move in the corresponding direction. \\
\bottomrule
\end{tabular}
\end{table*}

\clearpage

\subsection{Complete 2D and 3D Leaderboards}
\label{app:gamegen-subset-results}

We partition the 97 \GameGen{} tasks into 53 2D games and 44 3D games and
report the complete subset-level leaderboards in
\Cref{tab:gamegen-2d-results,tab:gamegen-3d-results}. All scores are reported
on a $[0,100]$ scale. Overall scores are computed from unrounded dimension
scores; consequently, averaging the displayed one-decimal values can differ
from the reported Overall score by 0.1 point.

\begin{table*}[t]
\centering
\caption{Results on the 53-game 2D subset of \GameGen{}. The best and
second-best results in each column are shown in bold and underlined,
respectively.}
\label{tab:gamegen-2d-results}
\small
\setlength{\tabcolsep}{5.5pt}
\begin{tabular}{clccccc}
\toprule
\# & \textbf{Model} & \textbf{Overall} & \textbf{Completeness} &
\textbf{Richness} & \textbf{Experience} & \textbf{Visual} \\
\midrule
1  & Claude-Opus-5     & \textbf{80.4} & \textbf{96.0} & \textbf{73.2} & \underline{73.6} & \textbf{78.8} \\
2  & Claude-Fable-5    & \underline{76.8} & \underline{91.6} & \underline{63.2} & \textbf{75.6} & \underline{77.2} \\
3  & Claude-Opus-4.8   & 71.0 & 86.4 & 52.0 & 71.2 & 74.0 \\
4  & Kimi-K3           & 70.5 & 84.0 & 54.4 & 70.4 & 73.2 \\
5  & Claude-Opus-4.7   & 67.9 & 84.8 & 49.2 & 67.6 & 70.0 \\
6  & DeepSeek-V4-Flash & 66.6 & 86.0 & 52.4 & 60.8 & 67.2 \\
7  & GLM5.2            & 65.7 & 79.2 & 46.4 & 65.2 & 72.0 \\
8  & GPT5.6-sol        & 65.2 & 80.0 & 44.0 & 67.2 & 69.6 \\
9  & Hy3         & 64.9 & 80.4 & 44.8 & 65.6 & 68.8 \\
10 & MiniMax-M3        & 64.0 & 82.0 & 43.6 & 60.0 & 70.4 \\
11 & Gemini-3.5-flash  & 62.2 & 77.6 & 49.2 & 56.4 & 65.6 \\
12 & GPT5.5            & 61.9 & 78.0 & 38.0 & 64.4 & 67.6 \\
13 & GLM5.1            & 58.9 & 73.2 & 45.2 & 57.6 & 59.2 \\
14 & Qwen3.7-Max       & 58.5 & 77.2 & 40.8 & 55.2 & 61.2 \\
15 & Seed-2.1-pro      & 54.2 & 66.4 & 34.8 & 51.6 & 64.0 \\
\bottomrule
\end{tabular}
\end{table*}

\begin{table*}[t]
\centering
\caption{Results on the 44-game 3D subset of \GameGen{}. The best and
second-best results in each column are shown in bold and underlined,
respectively.}
\label{tab:gamegen-3d-results}
\small
\setlength{\tabcolsep}{5.5pt}
\begin{tabular}{clccccc}
\toprule
\# & \textbf{Model} & \textbf{Overall} & \textbf{Completeness} &
\textbf{Richness} & \textbf{Experience} & \textbf{Visual} \\
\midrule
1  & Claude-Opus-5     & \textbf{78.9} & \textbf{92.0} & \textbf{70.4} & 70.8 & \textbf{82.4} \\
2  & Claude-Fable-5    & \underline{74.6} & 86.0 & \underline{62.4} & \textbf{73.2} & \underline{76.8} \\
3  & Kimi-K3           & 72.3 & \underline{87.2} & 58.0 & 70.0 & 74.0 \\
4  & Claude-Opus-4.8   & 68.7 & 81.6 & 46.4 & \underline{72.8} & 73.6 \\
5  & GPT5.6-sol        & 61.1 & 79.2 & 39.6 & 57.6 & 68.0 \\
6  & GLM5.2            & 60.9 & 74.0 & 43.2 & 59.2 & 66.8 \\
7  & Hy3         & 58.7 & 73.2 & 37.2 & 57.2 & 67.2 \\
8  & DeepSeek-V4-Flash & 58.5 & 76.8 & 43.2 & 53.6 & 60.4 \\
9  & GLM5.1            & 57.0 & 72.4 & 40.4 & 54.4 & 60.8 \\
10 & GPT5.5            & 55.5 & 68.8 & 35.2 & 55.6 & 62.4 \\
11 & Gemini-3.5-flash  & 55.2 & 68.0 & 42.0 & 48.8 & 62.4 \\
12 & Claude-Opus-4.7   & 55.0 & 65.6 & 35.2 & 54.4 & 64.8 \\
13 & MiniMax-M3        & 54.1 & 62.8 & 37.2 & 51.6 & 64.4 \\
14 & Qwen3.7-Max       & 49.4 & 54.0 & 26.4 & 55.2 & 62.4 \\
15 & Seed-2.1-pro      & 42.1 & 48.8 & 26.0 & 37.2 & 56.4 \\
\bottomrule
\end{tabular}
\end{table*}

\clearpage
\subsection{Per-Game Event Pass Rates}
\label{app:gamegen-pass-rates}

\begin{figure*}[p]
  \centering
  \includegraphics[height=0.86\textheight,keepaspectratio]{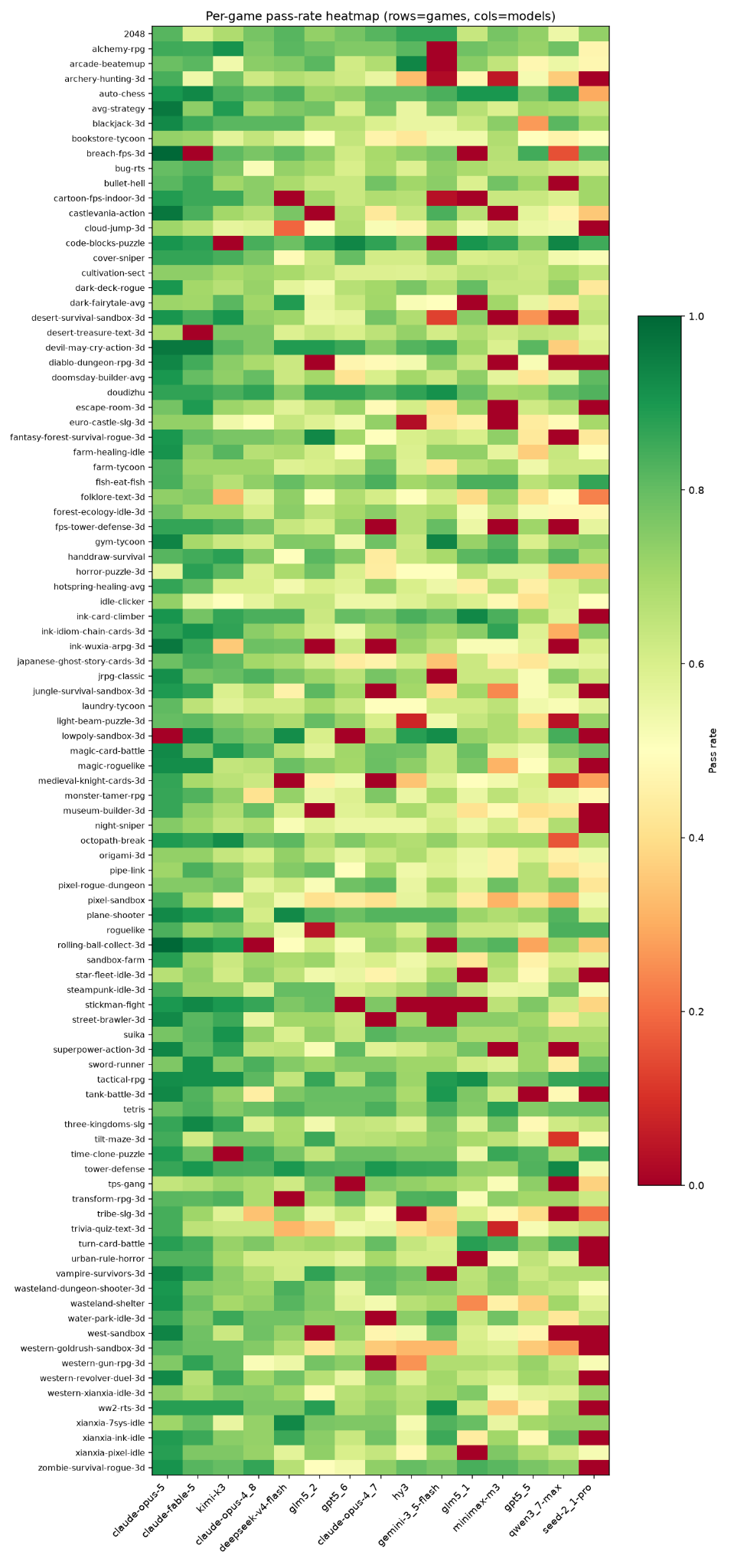}
  \caption{Per-game event pass rates across all 97 \GameGen{} tasks and 15
  evaluated models. Rows correspond to games, columns correspond to models,
  and each cell reports the fraction of evaluated checklist events that pass.
  Green indicates a higher pass rate and red indicates a lower pass rate.}
  \label{fig:gamegen-per-game-pass-rate}
\end{figure*}

\Cref{fig:gamegen-per-game-pass-rate} reports the event-level pass rate for
every game--model pair. The heatmap exposes substantial variation across games
that is hidden by aggregate model-level scores: even strong models encounter
isolated difficult games, while lower-ranked models occasionally perform well
on particular tasks.

%% file: appendices/2_gamefix_details.tex
\providecommand{\GameFix}{\textsc{GameFix}}
\providecommand{\strict}{\textsc{Strict}}
\providecommand{\general}{\textsc{General}}
\providecommand{\explicit}{\emph{Explicit Issue}}
\providecommand{\selfdisc}{\emph{Self-Discovery}}
\providecommand{\eg}{e.g.,}
\providecommand{\ie}{i.e.,}

\makeatletter
\@ifundefined{color@accentblue}{\definecolor{accentblue}{HTML}{1F4E96}}{}%
\@ifundefined{endcaseblock}{%
  \newenvironment{caseblock}[1]{\par\smallskip\noindent{\color{accentblue}\bfseries #1}\par\nobreak\vskip2pt\begingroup\small\setlength{\parskip}{3pt}}{\endgroup\par\smallskip}%
}{}
\@ifundefined{endmcard}{%
  \newenvironment{mcard}[1]{\par\medskip\noindent{\color{accentblue}\bfseries #1}\par\nobreak\vskip2pt\begingroup\small}{\endgroup\par\smallskip}%
}{}
\@ifundefined{enddeferbox}{%
  \newenvironment{deferbox}[1]{\par\smallskip\noindent{\bfseries #1}\par\nobreak\vskip2pt\begingroup\small\setlength{\parskip}{3pt}}{\endgroup\par\smallskip}%
}{}
\providecommand{\passpill}[1]{\textbf{[#1]}}
\providecommand{\failpill}[1]{\textbf{[#1]}}
\makeatother

\section{Additional Details of GameFix}
\label{app:gamefix}

This section expands \S\ref{sec:gamefix} with the F2P/P2P test definitions and multi-bug isolation (\S\ref{app:gf-tests}), the six agentic abilities measured by the benchmark (\S\ref{app:gf-abilities}), the complete secondary metrics for all 17 models (\S\ref{app:gf-full}), and the reasoning-level trajectory analysis (\S\ref{app:gf-traj}).

\subsection{F2P and P2P test definitions}
\label{app:gf-tests}

A test case is \emph{not} an equality check against the gold constant: it is a hand-written boolean predicate over the running game's observable state, expressed as a \textbf{behavioural tolerance band} (\eg a value must land in \texttt{(0, 1000]}, a displacement must exceed a threshold, or a velocity must have the correct sign). The gold patch is used only at \emph{authoring} time to certify that the tests fail on the mutated game, pass on the gold game, and remain stable across repeated gold runs. It plays no role when grading a candidate repair: any repair whose behaviour falls inside the intended band is credited.

\begin{itemize}[nosep]
\item \textbf{Fail-to-Pass (F2P)}: a predicate on the intended behaviour broken by the injected bug. It fails on the mutated game and must pass after a correct repair.
\item \textbf{Pass-to-Pass (P2P)}: a predicate on behaviour that must not regress. It passes on the gold game and must remain passing after repair, guarding against over-editing.
\end{itemize}

Each injected bug is counted as fixed only when its F2P test passes and all associated P2P tests remain passing. The task score is then the percentage of injected bugs that satisfy this criterion.

\paragraph{Multi-bug isolation: no cross-bug test contamination.}
\label{app:gf-iso}
Because each task co-injects 19--27 bugs into one game, a naive assertion set
admits two silent failure modes: a bug's F2P could be \emph{masked} green by a
neighbour's injection (a free pass for an unfixed bug), or its P2P could be
\emph{broken} by one (a spurious regression charged to an otherwise-correct
repair). We exclude both with an authoring-time gate. For each candidate task we
run three configurations of the game and require, per bug site $s$:
\begin{itemize}[nosep]
  \item \textbf{gold-green}: with the gold patch applied (all bugs fixed),
    every F2P of $s$ passes; the assertion is satisfiable.
  \item \textbf{coupled-base-red}: with all bugs co-injected (the base the
    candidate is judged on), $s$'s F2P is red; no neighbour masks it into a
    free pass.
  \item \textbf{single-bug-red}: with only $s$ injected, $s$'s F2P is red;
    the bug genuinely causes the symptom, not a merge artefact.
  \item \textbf{P2P survives the merged base}: $s$'s P2P passes on gold
    \emph{and} stays green on the fully-injected base; no co-injected neighbour
    can break it. A P2P any neighbour could break is rejected before shipping.
\end{itemize}
together with a stability check (gold $\to$ all P2P green for $N{=}5$
consecutive runs, zero flake). Bugs failing any check are dropped and replaced
from the pool under a line-disjointness constraint (no two bugs edit overlapping
lines, $\pm$ a window), and the gate re-runs until $\ge 20$ isolated bugs ship.
The four booleans are stored per task; \eg \texttt{avalanche-l5} records
\texttt{gold\_green = coupled\_base\_red = indep\_base\_red\_all = true} and
\texttt{p2p\_stable\_runs = 5}. This gate has a direct consequence at grading
time. Because the merged base is provably red on every F2P and
green on every P2P, a P2P failure on a candidate patch must be a regression
the candidate itself introduced, not contamination from an unfixed
neighbour, so the P2P gate also serves as our regression detector. On
\texttt{avalanche-l5}/\selfdisc{}, for instance, Hy3's fix to the cable-car
trigger drops a guard the gold patch keeps, so the \texttt{invuln-stun} P2P (``player
can move forward'') fails on a bug whose own F2P passes: a self-inflicted
regression, not a neighbour's contamination.

\subsection{The six agentic abilities the benchmark measures}
\label{app:gf-abilities}
The multi-bug, small-edit design (\S\ref{sec:gf-bugs}) does not test a single
``can it edit code'' ability. Each individual mutation is invisible in the
source and breaks only when the game is played, and a 19--27-bug task forces
several distinct abilities to act together. We separate the demanded skill into
six axes, shown in Table~\ref{tab:gf-abilities}, contrasting Hy3 (the
weakest model we analyse) with three stronger models, each with a verbatim
trajectory quote.

\begin{table}[h]
\centering
\footnotesize
\setlength{\tabcolsep}{5pt}
\renewcommand{\arraystretch}{1.15}
\caption{The six agentic ability axes, and how three models place on each. Each
cell carries a representative verbatim agent quote; quotes marked [tr.] are
translated from the agent's original Chinese.}
\label{tab:gf-abilities}
\begin{tabular}{@{}>{\raggedright\arraybackslash\hyphenpenalty=0 \exhyphenpenalty=0}p{0.125\textwidth}
                  >{\raggedright\arraybackslash\hyphenpenalty=0 \exhyphenpenalty=0}p{0.205\textwidth}
                  >{\raggedright\arraybackslash\hyphenpenalty=0 \exhyphenpenalty=0}p{0.175\textwidth}
                  >{\raggedright\arraybackslash\hyphenpenalty=0 \exhyphenpenalty=0}p{0.175\textwidth}
                  >{\raggedright\arraybackslash\hyphenpenalty=0 \exhyphenpenalty=0}p{0.165\textwidth}@{}}
\toprule
Axis & What it means & Claude Opus 5 & DeepSeek V4 Flash & Hy3 \\
\midrule
Self-discovery
& Find bugs the prompt never lists; recognise it is being tested
& ``what the original\ldots had to understand which bugs were \emph{intentionally introduced}''
& ``this is a `find the sabotage' task\ldots a subtle modification from the original''
& ``All $13{+}2$ verified'' yet 1/22: \emph{does not search} \\
\addlinespace[2pt]
Behavioural verification
& Games are stateful and timed; verify by running the game
& ``total failures: 0/60''; runs the physics to verify fixes
& self-authored harness; fix\discretionary{}{}{}$\to$\discretionary{}{}{}run\discretionary{}{}{}$\to$\discretionary{}{}{}observe loop
& ``I couldn't runtime-\hspace{0pt}playtest''; ``environmental, not a code issue'' \\
\addlinespace[2pt]
Value recovery
& Recover constants defined by game feel, not text (gravity, radius, jump)
& derives from geometry: ``DEFAULT\_\hspace{0pt}DAMPING\_\hspace{0pt}X\ldots so top speed equals Player.\hspace{0pt}SPEED''
& re-invents: ``what makes a PLAYABLE game'' (often misses gold)
& guesses but does not verify: ``simulation confirms the fix works'' [tr.] (6/22) \\
\addlinespace[2pt]
Multi-bug coverage \& planning
& Localise, prioritise, chase prerequisite chains across 19--27 bugs
& near-perfect under \explicit{}; no early stop
& strong but over-edits (79 edits)
& early stop: finds \texttt{introTime++} root cause, fixes only 6, leaves 13 \\
\addlinespace[2pt]
Regression control
& Don't break adjacent behaviour while fixing
& near-zero regression
& 0.18 regressions/cell
& 0.50/cell, 8/50 cells; fixes 13 symptoms but regresses 12 $\to$ 1/22 \\
\addlinespace[2pt]
Stopping criterion
& Decide when a task is complete
& objective: external diff ``harness fixed exactly the 19\ldots confirms injected issues''
& plausibility: ``8 it is'' (admits gamble)
& subjective + over-conservative: ``I deliberately left this alone\ldots it is not among the five items you listed'' [tr.] \\
\bottomrule
\end{tabular}
\end{table}

Two points about the table. First, the axes are not independent:
\emph{behavioural verification} and \emph{regression control} are two sides of
whether the model runs the game to completion, the first being the ability to
build a real test harness and the second being what that harness then catches.
Second, five of the six axes appear mainly under \selfdisc{}. Under \explicit{}
the symptom list supplies the localisation, the completeness target, and (per
\S\ref{app:gf-iso}) a guard against silent regression, so \explicit{} collapses
several axes into ``follow the list.'' Removing the list, as \selfdisc{} does,
turns each axis into a separate, discriminating demand, so the
\explicit{}$\to$\selfdisc{} cliff measures the six axes rather than a single
coding skill.

\subsection{Full leaderboard with secondary metrics}
\label{app:gf-full}

Table~\ref{tab:gf-full} reports, for all 17 models: pass@3 (any of 3 runs solves
the task), pass$^3$ (all 3 runs solve it) and average@3 counts at score
thresholds 100/95/90; mean agent turns per clean-completed task; and mean
per-task uncached input / output tokens. Blank token/turn tasks indicate a
framework whose usage stream did not expose the field. All are derived from the
same live report module as the main leaderboard.

\begin{table}[h]
  \centering
    \caption{Secondary metrics for all 17 models. Counts are out of 100 tasks
  (pass@3, pass$^3$) or expected tasks (avg@3). Models evaluated under two agent
  frameworks are listed once per framework, given in parentheses; all other
  models are run under Claude Code. The Hy3 figures are from the
  self-deployed model (see \S\ref{sec:gf-results}).}
  \label{tab:gf-full}
  \footnotesize
  \setlength{\tabcolsep}{8.5pt}
  \renewcommand{\arraystretch}{1.05}
  \begin{tabular*}{\textwidth}{@{\extracolsep{\fill}}l l rrr rrr rrr@{}}
    \toprule
    & & \multicolumn{3}{c}{pass@3 ($\ge$)} & \multicolumn{3}{c}{pass$^3$ ($\ge$)} & \multicolumn{3}{c}{avg@3 ($\ge$)} \\
    \cmidrule(lr){3-5} \cmidrule(lr){6-8} \cmidrule(lr){9-11}
    Model & Effort & 100 & 95 & 90 & 100 & 95 & 90 & 100 & 95 & 90 \\
    \midrule
    Claude Opus 5                 & max   & 31 & 56 & 76 & 14 & 35 & 60 & 22.3 & 44.7 & 68.3 \\
    Claude Fable 5                & max   & 26 & 52 & 69 & 10 & 29 & 49 & 18.0 & 40.0 & 60.3 \\
    GPT-5.6-sol (Codex)           & xhigh & 22 & 44 & 69 & 13 & 23 & 42 & 17.0 & 32.7 & 55.0 \\
    GPT-5.6-sol (Claude Code)     & xhigh & 24 & 43 & 68 & 6  & 18 & 35 & 14.0 & 30.7 & 51.0 \\
    Claude Opus 4.8               & max   & 11 & 31 & 45 & 3  & 16 & 25 & 7.0  & 23.3 & 34.3 \\
    Claude Opus 4.7               & max   & 12 & 31 & 48 & 1  & 15 & 22 & 5.7  & 22.3 & 36.0 \\
    GPT-5.5 (Codex)               & xhigh & 14 & 27 & 54 & 2  & 10 & 23 & 7.3  & 18.7 & 38.3 \\
    GPT-5.5 (Claude Code)         & xhigh & 13 & 28 & 56 & 2  & 8  & 22 & 6.7  & 18.7 & 39.0 \\
    DeepSeek V4 Flash             & max   & 11 & 32 & 46 & 0  & 9  & 20 & 5.0  & 20.7 & 32.3 \\
    Kimi K3                       & max   & 8  & 26 & 45 & 2  & 10 & 20 & 5.0  & 17.0 & 32.7 \\
    GLM 5.2                       & xhigh & 9  & 26 & 40 & 2  & 7  & 17 & 5.3  & 16.0 & 29.3 \\
    Gemini 3.5 Flash              & high  & 10 & 21 & 39 & 3  & 7  & 18 & 5.7  & 13.7 & 28.3 \\
    MiniMax-M3                    & on    & 10 & 25 & 35 & 0  & 2  & 9  & 4.3  & 13.0 & 21.3 \\
    Seed-2.1-pro                  & high  & 5  & 22 & 33 & 0  & 3  & 8  & 2.0  & 10.7 & 19.7 \\
    GLM 5.1                       & xhigh & 2  & 16 & 30 & 0  & 3  & 9  & 1.0  & 8.7  & 20.0 \\
    Hy3                           & high  & 3  & 12 & 22 & 0  & 2  & 9  & 1.3  & 7.0  & 16.0 \\
    Qwen3.7-Max                   & max   & 2  & 13 & 26 & 0  & 1  & 4  & 0.7  & 7.3  & 15.3 \\
    \bottomrule
  \end{tabular*}
\end{table}

\subsection{Reasoning-level trajectory analysis}
\label{app:gf-traj}

We analyse agent trajectories only within the information exposed by each model provider and framework. Frontier models generally do not provide raw chain-of-thought traces, so our analysis does not rely on hidden reasoning. Instead, we examine the available reasoning summaries or planning preambles, the model's visible outputs, tool-use sequences, code edits, and final task outcomes. For each representative model, we inspect these observable signals across selected tasks and relate failure cases to the benchmark taxonomy. This allows us to compare search, repair, verification, and stopping behaviour without requiring access to private chain-of-thought. Table~\ref{tab:gf-traj-summary} summarises the dominant failure mechanism for each analysed model.

\begin{table}[h]
  \centering
  \small
  \setlength{\tabcolsep}{4.5pt}
  \renewcommand{\arraystretch}{1.18}
  \begin{tabular*}{\textwidth}{@{\extracolsep{\fill}} l l r r >{\raggedright\arraybackslash}p{0.34\textwidth} @{}}
    \toprule
    Model & Profile & \strict{} & Cliff & Dominant cliff mechanism \\
    \midrule
    Claude Opus 5     & forensic auditor      & 39.0        & 7.6  & no dominant failure mode: self-discovery is the default; small losses are mainly scope-related \\
    Claude Opus 4.8   & throttled auditor     & 18.0        & 17.9 & fix-authorization: finds additional bugs but sometimes declines to repair them \\
    GPT-5.6-sol       & framework control     & 29.1 / 26.7 & 11.3 & hint-anchored stopping: often stops after covering the disclosed issues; framework effect is small \\
    DeepSeek V4 Flash & behavioural repairer  & 15.2        & 13.6 & limited search coverage: verifies repairs well but does not always discover all hidden bugs \\
    Kimi K3           & best open model       & 14.0        & 19.6 & missed hidden bugs and unreliable self-verification \\
    GLM 5.2           & disciplined restorer  & 13.4        & 20.2 & scope-dependent fixing: often declines unlisted bugs when intent is ambiguous \\
    Hy3               & checklist-dependent   & 6.0         & 32.8 & discovery deficit and premature stopping \\
    \bottomrule
  \end{tabular*}
  \caption{The seven trajectory-analysed models (marked $^{\dagger}$ in
  Table~\ref{tab:gf-leaderboard}), their profile and dominant cliff mechanism.
  GPT-5.6-sol \strict{} is shown as Codex / Claude Code. Cliff is the drop from
  \explicit{} to \selfdisc{}; all values 3-run average@3.}
  \label{tab:gf-traj-summary}
\end{table}

%% file: appendices/3_gameopt_details.tex
\section{\GameOpt{} Details}
\label{app:gameopt}

This appendix provides supplementary details for the 17-chain JavaScript
collection evaluated in \S\ref{sec:gameopt-results}.  We first summarize
the chain inventory and rubric composition, then provide three request chains
and one complete hidden rubric.  Each starting snapshot runs in the offline
browser sandbox, so acceptance criteria may require rendered output as
evidence.

\subsection{Track Overview}
\label{app:gameopt-overview}

\begin{table}[H]
  \centering
  \small
  \setlength{\tabcolsep}{4.5pt}
  \renewcommand{\arraystretch}{1.08}
  \caption{JavaScript chain inventory.}
  \label{tab:js-chains}
  \begin{tabular}{@{}llrrrrrr@{}}
    \toprule
    & & \multicolumn{2}{c}{\textbf{Snapshot}} & & \multicolumn{3}{c}{\textbf{Criteria}} \\
    \cmidrule(lr){3-4}\cmidrule(l){6-8}
    \textbf{Chain} & \textbf{Start} & \textbf{Files} & \textbf{kchar} & \textbf{Diff.} & \textbf{Total} & \textbf{P0} & \textbf{Reg.} \\
    \midrule
    Basketball & first version & 12 & 271 & 5 & 42 & 6 & 6 \\
    Deck-Climb Roguelike & \texttt{v0004} & 9 & 76 & 3 & 43 & 7 & 7 \\
    Deck-Climb Roguelike B & \texttt{v0001} & 4 & 72 & 4 & 40 & 6 & 6 \\
    Hero Arena & \texttt{v0425} & 49 & 546 & 3 & 40 & 8 & 7 \\
    Hot Potato & first version & 9 & 48 & 5 & 42 & 7 & 6 \\
    Kart Racer & first version & 20 & 272 & 5 & 42 & 6 & 6 \\
    Lane Battler & \texttt{v0001} & 9 & 106 & 4 & 40 & 6 & 6 \\
    Marigold Dash & first version & 13 & 111 & 5 & 42 & 6 & 6 \\
    Mech Arena & \texttt{v0004} & 15 & 121 & 4 & 40 & 6 & 7 \\
    Mythic Boss Fight & first version & 33 & 366 & 5 & 42 & 6 & 6 \\
    Neon Runner & \texttt{v0002} & 3 & 83 & 4 & 41 & 6 & 6 \\
    Offering Scramble & first version & 13 & 101 & 5 & 42 & 6 & 6 \\
    Open-World Adventure & first version & 12 & 203 & 5 & 42 & 6 & 6 \\
    Paper-Cut Stand & first version & 12 & 77 & 5 & 42 & 6 & 6 \\
    Pool Master & \texttt{v0001} & 2 & 50 & 3 & 40 & 6 & 4 \\
    Street Racer & \texttt{v0001} & 9 & 92 & 3 & 39 & 6 & 0 \\
    Tactical FPS & first version & 18 & 304 & 5 & 42 & 6 & 6 \\
    \midrule
    \textbf{Total} (17 chains) & & \textbf{242} & \textbf{2,898} & & \textbf{701} & \textbf{106} & \textbf{97} \\
    \bottomrule
  \end{tabular}
\end{table}

Table~\ref{tab:js-chains} covers 17 chains, 102 turns, and 701 criteria.
Snapshot size counts authored source in $G^{(0)}$, excluding vendored engine
builds and \texttt{node\_modules}; difficulty is rated on a 1--5 scale.  A
\emph{first version} start denotes the first recorded version,
whereas other rows use the named version from a co-creation history.  The
intended budget is one P0 and one regression check per turn.  Fourteen chains
meet the P0 budget exactly, but \emph{Street Racer} has no regression check.
Three representative chains are expanded in Table~\ref{tab:js-showcase}.

\begin{table}[H]
  \centering
  \small
  \setlength{\tabcolsep}{8pt}
  \renewcommand{\arraystretch}{1.08}
  \caption{Acceptance-criterion composition.}
  \label{tab:js-composition}
  \begin{tabular}{@{}lrr@{}}
    \toprule
    \textbf{Group} & \textbf{Count} & \textbf{Share (\%)} \\
    \midrule
    \multicolumn{3}{@{}l}{\textit{Criterion category}} \\
    \addlinespace[2pt]
    \quad requirement & 392 & 55.9 \\
    \quad challenge & 212 & 30.2 \\
    \quad regression & 97 & 13.8 \\
    \addlinespace[4pt]
    \multicolumn{3}{@{}l}{\textit{Admissible evidence}} \\
    \addlinespace[2pt]
    \quad code only & 281 & 40.1 \\
    \quad rendered output only & 298 & 42.5 \\
    \quad code $+$ rendered output & 122 & 17.4 \\
    \addlinespace[4pt]
    \multicolumn{3}{@{}l}{\textit{Provenance}} \\
    \addlinespace[2pt]
    \quad \texttt{user\_prompt} & 546 & 77.9 \\
    \quad \texttt{engineering} & 86 & 12.3 \\
    \quad \texttt{diff\_analysis} & 61 & 8.7 \\
    \quad \texttt{real\_user\_followup} & 3 & 0.4 \\
    \quad \texttt{synthetic} & 5 & 0.7 \\
    \addlinespace[4pt]
    \multicolumn{3}{@{}l}{\textit{Other criterion properties}} \\
    \addlinespace[2pt]
    \quad proxy & 33 & 4.7 \\
    \quad regression: major / minor & 48 / 49 & 13.8 \\
    \bottomrule
  \end{tabular}
\end{table}

Percentages in Table~\ref{tab:js-composition} use all criteria as the
denominator.
\texttt{diff\_analysis} criteria come from changes between adjacent human
versions; \texttt{real\_user\_followup} criteria restate later complaints from
the same user.

\begin{table}[H]
  \centering
  \small
  \setlength{\tabcolsep}{9pt}
  \renewcommand{\arraystretch}{1.08}
  \caption{Dimension order by turn.}
  \label{tab:js-dimension-order}
  \begin{tabular}{@{}lrrrrrr@{}}
    \toprule
    & \multicolumn{6}{c}{\textbf{Turn}} \\
    \cmidrule(l){2-7}
    \textbf{Dimension} & \textbf{1} & \textbf{2} & \textbf{3} & \textbf{4} & \textbf{5} & \textbf{6} \\
    \midrule
    Gameplay & 10 & 2 & 4 & 1 & -- & -- \\
    Level & -- & 10 & -- & 4 & 3 & -- \\
    Balance & 2 & 2 & 11 & 1 & 1 & -- \\
    Art & 1 & 1 & 2 & 9 & 4 & -- \\
    UI & 4 & 2 & -- & 2 & 9 & -- \\
    Audio & -- & -- & -- & -- & -- & 17 \\
    \bottomrule
  \end{tabular}
\end{table}

Each chain covers every dimension exactly once.  The ordering is only
partially counterbalanced: \textsc{audio} is always turn six, so turn position
and dimension cannot be fully separated, as noted in
\S\ref{sec:gameopt-results}.

\subsection{Request Chains}
\label{app:gameopt-chains}

The requests below are translated from the Chinese originals shown to the
model.  The translations preserve register and deliberate vagueness; resolving
phrases such as ``make these boundaries more clearly visible'' is part of the
task.

\clearpage
\begingroup
\footnotesize
\setlength{\tabcolsep}{3pt}
\setlength{\LTpre}{6pt}
\setlength{\LTpost}{10pt}
\renewcommand{\arraystretch}{1.12}
\begin{longtable}{@{}>{\centering\arraybackslash}p{0.04\textwidth}
                      >{\raggedright\arraybackslash}p{0.11\textwidth}
                      >{\raggedright\arraybackslash}p{0.62\textwidth}
                      >{\centering\arraybackslash}p{0.04\textwidth}
                      >{\centering\arraybackslash}p{0.035\textwidth}
                      >{\centering\arraybackslash}p{0.045\textwidth}@{}}
  \caption{Representative request chains.}
  \label{tab:js-showcase}\\
  \toprule
  \textbf{\#} & \textbf{Dimension} & \textbf{Request shown to the model} & \textbf{N} & \textbf{P0} & \textbf{Img.} \\
  \midrule
  \endfirsthead
  \multicolumn{6}{c}{\small\tablename~\thetable{} (continued)} \\
  \addlinespace[2pt]
  \toprule
  \textbf{\#} & \textbf{Dimension} & \textbf{Request shown to the model} & \textbf{N} & \textbf{P0} & \textbf{Img.} \\
  \midrule
  \endhead
  \midrule
  \multicolumn{6}{r}{\textit{Continued on the next page}} \\
  \endfoot
  \bottomrule
  \endlastfoot
    \multicolumn{6}{@{}l}{\textbf{Street Racer} --- \texttt{v0001}; 9 files; 92 kchar; difficulty 3; 39 criteria} \\
    \addlinespace[2pt]
    1 & UI & Unify the whole game's UI into a dark-background street-arcade look: near-black translucent panels, bright-yellow primary buttons and outlines, one self-drawn icon set and colour code (green = score / steady, orange = combo / heat, white = time and speed digits). Replace the default system controls, and keep information at the screen edges so it does not block the driving view. Cover the main menu, the car-select page, the in-game HUD and the results page. Core requirements: (1) all four screens share the dark arcade style and self-drawn controls, with a clear primary/secondary button hierarchy; (2) structure the in-game HUD --- minimap and settings entry top-left, wanted/heat bar, score and time (mm:ss) top-centre, combo plate top-right, a virtual stick bottom-left that actually drives the car, speed digits plus handbrake/reset/horn buttons bottom-right; (3) complete the feedback --- floating score on smashing props, a large combo readout whose multiplier changes colour, and a score tick. The results page presents the run, highlights a new record, and uses the same primary/secondary buttons. & 8 & 1 & 8 \\
    2 & Gameplay & While driving, the car can enter or even pass straight through walls over a large area. Please prevent this: the car should not be able to get inside walls or buildings. & 6 & 1 & 5 \\
    3 & Art & In some places there are invisible air walls. I would like these boundaries to be delimited and made more clearly visible. & 6 & 1 & 6 \\
    4 & Level & On the screen where the game starts, let the player freely choose the map, and add more maps to choose from. & 7 & 1 & 7 \\
    5 & Balance & Among the six cars, the supercar ``Ghost X'' has nearly the highest top speed, acceleration and handling --- it is an all-rounder, so there is no reason to pick any of the others. Please rebalance the cars' numbers: give every car a clear strength and a clear weakness so that each has a distinct role. Also make the ``weight'' attribute actually affect how driving feels --- for instance a heavy car is more stable and takes hits better but steers more sluggishly, while a light car is more agile but is knocked around more easily. Do not let any single car be the best on every attribute. & 6 & 1 & 1 \\
    6 & Audio & The sound effects feel good overall, but there are two problems I want solved. First, there is nowhere to turn the sound off --- after playing a while with headphones I cannot lower the volume or mute, so I want a volume/mute control in the pause panel, and the setting should be remembered. Second, when I smash a whole row of props the breaking sounds pile up in an instant into a loud crackling mush; please rein in this kind of high-frequency effect so they do not all cram into the same moment. & 6 & 1 & 1 \\
    \addlinespace[4pt]
    \midrule
    \multicolumn{6}{@{}l}{\textbf{Deck-Climb Roguelike} --- \texttt{v0004}; 9 files; 76 kchar; difficulty 3; 43 criteria} \\
    \addlinespace[2pt]
    1 & Balance & Adjust the game balance to the following number system: starting health 70, 3 energy per turn, a base hand of 5 cards; a basic attack costs 1 energy and deals about 6 damage, a defence card grants about 5 shield; normal enemies have 30--50 health, elites 80--120, and the boss about 200--300; after a fight the player picks 1 card out of 3, and may skip so the deck does not get too thick; the shop can sell cards and relics or remove a card, and a rest site restores about 30\% of max health. & 8 & 1 & 4 \\
    2 & Level & Change the levels into a vertical branching tower-climb map with combat, elite, event, rest, shop and treasure nodes. Keep early enemy mechanics simple, introduce status ailments and multi-enemy fights in the middle, and use elites and the boss late to test the strength of the deck; different routes should carry different risk and reward. & 8 & 1 & 5 \\
    3 & Gameplay & The cards are currently not strategic enough to beat the final boss. Please improve three things: first, there is no way to heal during a fight, so add one; second, the boss has too much health, so lower it somewhat; third, there are too few strategic card types --- for example there is no ``dodge'' card that completely avoids the damage of one enemy attack, so add cards of that kind. & 7 & 1 & 4 \\
    4 & UI & Please add deck editing and a card compendium. & 7 & 1 & 5 \\
    5 & Art & Rework the UI into the same fresh hand-drawn campus style as the game: panels, buttons and cards should systematically carry a hand-drawn quality (outlines / texture / a rounded sticker feel) across the main menu, the map, the combat HUD and the results screen, staying with the existing warm cream / grass / red-blue-yellow palette. While you are at it, fix the places where controls overlap or sit too close together and end up occluding key information (health, energy, gold, enemy health bars, the hand). & 7 & 2 & 5 \\
    6 & Audio & The sound is a bit muddy right now: many actions sound the same --- playing a card, taking a hit, getting a debuff and losing are often the same sound --- and drawing a card makes no sound at all. On top of that, when I play cards quickly in a row, or a pile of enemies acts at once, the sounds all stack up and it gets very loud, almost clipping. Could you separate the sounds for these key actions, fill in the ones that are silent, and give me a mute/volume control? & 6 & 1 & 1 \\
    \addlinespace[4pt]
    \midrule
    \multicolumn{6}{@{}l}{\textbf{Hero Arena} --- \texttt{v0425}; 49 files; 546 kchar; difficulty 3; 40 criteria} \\
    \addlinespace[2pt]
    1 & Balance & The talent level cap is 20. Please make this cap explicit and enforce it in the system. & 6 & 1 & 4 \\
    2 & Gameplay & When each wave's timer ends, first bring every character and monster in the game to a halt; then delete all the monsters' sprite frames and entities, and only enter the shop screen once the deletion has finished. The deletion can be done with a spin-and-shrink tween. & 7 & 1 & 7 \\
    3 & Art & Add suitable sound effects for button presses, ranged weapon fire, melee weapon swings and thrusts, and shell explosions. & 7 & 3 & 0 \\
    4 & UI & Change the way the Extreme difficulty is entered: clicking the Extreme option should no longer go straight into the game, but show an Extreme leaderboard first, with a Start button below it; only pressing Start enters the Extreme game scene. & 7 & 1 & 5 \\
    5 & Level & Monster spawn positions in the arena are too random --- several groups in a row often come from the same direction, so the player only has to keep moving the other way. Please stagger each group's spawn direction around the player so monsters close in from different sides; at the same time keep the existing no-spawn-in-your-face distance, and make sure spawn points stay inside the arena bounds. & 6 & 1 & 3 \\
    6 & Audio & From the main town to combat and on to the victory/defeat results, the background music is one and the same loop --- the mood never changes; and the moment I win or lose it is dead silent. I want combat to have its own battle music, and the victory/defeat results screen to switch to a matching victory or defeat track (or sting). Also, when health is nearly gone, give me a continuous low-health warning sound so I know I am about to die. Do not break the existing fire, hit and button sounds. & 7 & 1 & 0 \\
\end{longtable}
\endgroup

Table~\ref{tab:js-showcase} reports all six requests for each example, along
with the total, P0, and rendered-output criterion counts.  The examples span
interface, gameplay, level design, balance, art, and audio optimization.
\emph{Deck-Climb Roguelike} turn 1 is a fully quantified balance request, while
\emph{Hero Arena} turn 3 is expanded below.

\clearpage
\subsection{Rubric Example}
\label{app:gameopt-rubric}

Table~\ref{tab:js-rubric-example} gives the complete hidden rubric for
\emph{Hero Arena}, turn 3.  Its visible request asks only for suitable sound
effects for buttons, ranged fire, two melee attacks, and shell explosions.  All
seven items use code evidence, so the table omits a redundant evidence column.

\begin{table}[H]
  \centering
  \footnotesize
  \setlength{\tabcolsep}{3pt}
  \renewcommand{\arraystretch}{1.12}
  \caption{Example hidden rubric: \emph{Hero Arena}, turn 3.}
  \label{tab:js-rubric-example}
  \begin{tabular}{@{}>{\raggedright\arraybackslash}p{0.035\textwidth}
                      >{\raggedright\arraybackslash}p{0.16\textwidth}
                      >{\raggedright\arraybackslash}p{0.60\textwidth}
                      >{\raggedright\arraybackslash}p{0.14\textwidth}@{}}
    \toprule
    \textbf{ID} & \textbf{Type} & \textbf{Condition} & \textbf{Source} \\
    \midrule
    R1 & Req. / P0 & Clickable in-game buttons play a click sound on press, and the coverage includes the start-screen / main-screen buttons, not only the in-combat UI buttons. & user follow-up \\
    R2 & Req. / P0 & Firing a ranged weapon plays a fire sound (the sound may be procedurally synthesised with WebAudio or reuse an existing source; a real recorded asset is not required). & user prompt \\
    R3 & Req. / P1 & A melee attack issues a sound call at the attack trigger point, and the two melee types (\texttt{meleeType} swing vs.\ thrust, or an equivalent branch) use different sound parameters: the sound name or synthesis parameters (waveform, \texttt{freq}/\texttt{freqEnd}, noise, \ldots) differ between the two calls rather than sharing one setting. & user prompt \\
    R4 & Req. / P1 & A shell with an explosion effect plays an explosion sound when it detonates (procedural synthesis is acceptable). & user prompt \\
    R5 & Req. / P0 & [proxy] On the code path of each trigger point (button \texttt{pointerdown}, fire, melee attack, shell explosion) there is a call that actually reaches the sound interface (e.g.\ \texttt{AudioManager.playSfx}/\texttt{playSfxEx} or \texttt{ProceduralSfx.play}/\texttt{playTone}) and that call actually reaches audio output (a WebAudio oscillator/noise node or an existing source) --- not merely a config entry or constant with no playback call anywhere. The criterion is the existence of the trigger-to-sound-call binding (the real user reported hearing no sound at all). & user follow-up \\
    R6 & Challenge / P1 & High-frequency events (sustained fire, melee combos) have demonstrable rate limiting in code: the sound call passes a minimum-interval or throttle parameter (e.g.\ \texttt{throttle\_ms}/\texttt{throttleMs} or an equivalent interval test) or a concurrency cap, rather than playing unconditionally on every trigger. & diff analysis \\
    R7 & Regression / minor, $-2$ & Does not break existing behaviour: in the final code the hit, kill, damage, gold-pickup and purchase sounds already present in the starting snapshot still fire as they did in the snapshot. & engineering \\
    \bottomrule
  \end{tabular}
\end{table}

Three of the seven criteria are P0.  Two of them (\texttt{R1} and
\texttt{R5}) come from \texttt{real\_user\_followup}: later reports that the
buttons remained silent and that no output was audible motivate an explicit
trigger-to-output condition.  \texttt{R6} adds throttling from the adjacent
human-version diff, although the request does not mention it.  \texttt{R7}
awards no positive credit and can only subtract.  This example illustrates the
provenance result in \S\ref{sec:gameopt-results}: satisfying the stated request
is only the first layer of evaluation.